%% file: main.tex
\documentclass[10pt,twocolumn,letterpaper]{article}

\usepackage{cvpr}      


\usepackage{subcaption}
\usepackage{graphicx}

\usepackage{multirow}
\usepackage{multicol}
\usepackage{fontawesome5}
\usepackage[accsupp]{axessibility}
\usepackage{fontawesome5}
\usepackage{pdflscape}
\usepackage{pict2e,xcolor}

\newcommand\mdoubleplus{\mathbin{+\mkern-10mu+}}

\newcommand{\fplus}[1][black]{%
  \begingroup\leavevmode\color{#1}%
  \setlength{\unitlength}{1em}%
  \linethickness{.33em}%
  \begin{picture}(1,1)
  \put(0,0.5){\line(1,0){1}}
  \put(0.5,0){\line(0,1){1}}
  \end{picture}%
  \endgroup
}

\definecolor{cvprblue}{rgb}{0.21,0.49,0.74}
\usepackage[pagebackref,breaklinks,colorlinks,allcolors=cvprblue]{hyperref}

\def\paperID{*****} 
\def\confName{CVPR}
\def\confYear{2025}

\title{OmniGCD: Abstracting Generalized Category Discovery for Modality Agnosticism}

\author{
Jordan Shipard$^{1,2}$\thanks{Work was done during PhD at SAIVT, QUT.} \quad
Arnold Wiliem$^{1,2}$ \quad
Kien Nguyen Thanh$^{1}$ \quad
Wei Xiang$^{3}$ \quad
Clinton Fookes$^{1}$\\[0.3em]
$^{1}$SAIVT, QUT, Australia \quad
$^{2}$Shield AI, Australia \quad
$^{3}$La Trobe University, Australia\\[0.3em]}

\begin{document}
\maketitle
\begingroup
\renewcommand\thefootnote{}
\footnotetext{Emails: \{jordan.shipard, arnold.wiliem\}@shield.ai,  \{k.nguyenthanh, c.fookes\}@qut.edu.au, w.xiang@latrobe.edu.au}
\endgroup
\input{sec/0_abstract}   
\input{sec/1_intro}
\input{sec/2_related_works}
\input{sec/3_method}

\input{sec/4_results}
\input{sec/5_ablations}
\input{sec/7_conclusion}


{
    \small
    \bibliographystyle{ieeenat_fullname}
    \bibliography{main}
}



\twocolumn[
\begin{center}
{\LARGE \textbf{Supplementary Material for ``OmniGCD: Abstracting Generalized Category Discovery for Modality Agnosticism"}}\par
\vspace{1em}
\end{center}
]
\input{supp_sec/supp_intro}
\input{supp_sec/GCDFormer_optim}
\input{supp_sec/dataset_stats}
\input{supp_sec/OmniGCD_dim_ablation}
\input{supp_sec/additional_encoders}
\input{supp_sec/tsne_vs_gcdformer}
\input{supp_sec/finetuning}
\input{supp_sec/dim_reduction_results}

\end{document}

%% file: sec/0_abstract.tex
\begin{abstract}
Generalized Category Discovery (GCD) challenges methods to identify known and novel classes using partially labeled data, mirroring human category learning. Unlike prior GCD methods, which operate within a single modality and require dataset-specific fine-tuning, we propose a modality-agnostic GCD approach inspired by the human brain's abstract category formation. Our \textbf{OmniGCD} leverages modality-specific encoders (e.g., vision, audio, text, remote sensing) to process inputs, followed by dimension reduction to construct a \textbf{GCD latent space}, which is transformed at test-time into a representation better suited for clustering using a novel synthetically trained Transformer-based model. To evaluate OmniGCD, we introduce a \textbf{zero-shot GCD setting} where no dataset-specific fine-tuning is allowed, enabling modality-agnostic category discovery. \textbf{Trained once on synthetic data}, OmniGCD performs zero-shot GCD across 16 datasets spanning four modalities, improving classification accuracy for known and novel classes over baselines (average percentage point improvement of \textbf{+6.2}, \textbf{+17.9}, \textbf{+1.5} and \textbf{+12.7} for vision, text, audio and remote sensing). This highlights the importance of strong encoders while decoupling representation learning from category discovery. Improving modality-agnostic methods will propagate across modalities, enabling encoder development independent of GCD. Our work serves as a benchmark for future modality-agnostic GCD works, paving the way for scalable, human-inspired category discovery. All code is available at \href{https://github.com/Jordan-HS/OmniGCD}{github.com/Jordan-HS/OmniGCD}. 
\end{abstract}

%% file: sec/1_intro.tex

\vspace{-15pt}
\section{Introduction}

\begin{figure}
    \centering
    \includegraphics[width=\linewidth]{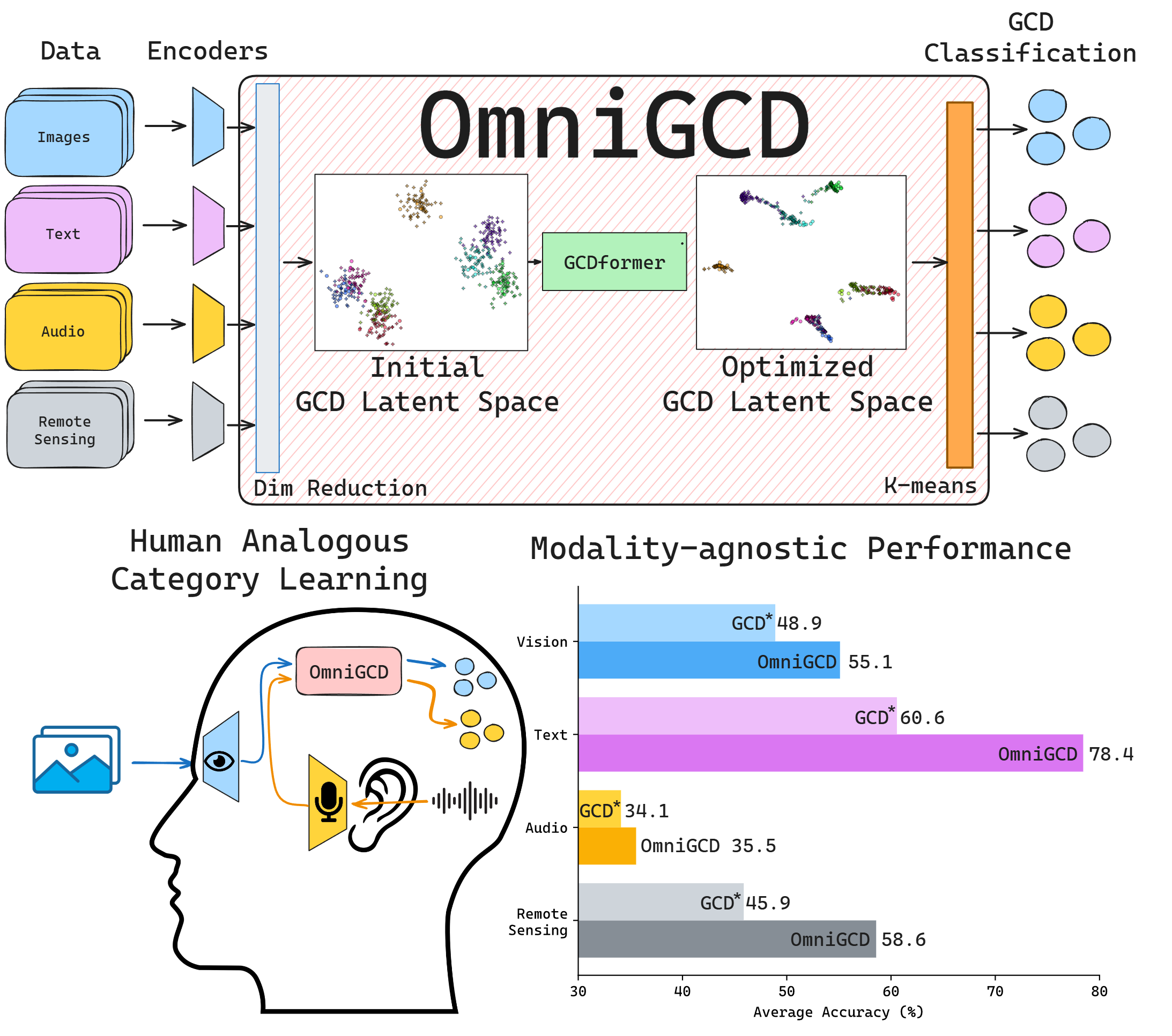}
    \caption{Summary of OmniGCD for modality-agnostic zero-shot GCD, inspired by human category learning. OmniGCD utilizes modality-specific encoders to construct a GCD latent space, which is transformed at test-time into a representation better suited for clustering by GCDformer. $GCD^*$ refers to the original GCD method~\cite{GCD2022} without fine-tuning. 
    }
    \label{fig:summary}
\end{figure}

Inspired by an infant observing and learning from the world around it, Generalized Category Discovery (GCD)~\cite{GCD2022} was proposed as a more realistic and human centric computer vision task, compared to traditional classification. It adds the fundamental question, do we recognize what we are observing, or is it novel? In neuroscience, this ability to form common categories is referred to as \textit{category learning}~\cite{ashby2005human}. It is understood that in the human brain, the prefrontal cortex integrates highly preprocessed signals from other major forebrain areas, and is responsible for higher-order reasoning tasks~\cite{miller2001integrative}, among which is the ability to form categories~\cite{miller2002prefrontal, freedman2001categorical}. Additionally, recent studies have shown that category formation and representation is independent of specific sensory inputs~\cite{jung2018modality, dirani2024meg}. This shows that category formation is fundamentally an abstract task, independent of any specific modality. To our knowledge, all GCD works investigate category discovery within an isolated domain, such as vision~\cite{tang2025dissecting, muGCD23} or text~\cite{loop, glean}, and require independent fine-tuning on each dataset they are tested on. While highly effective within this framework, these methods ignore the fundamental abstraction of category learning that is demonstrated in humans. This raises the question of whether current methods truly achieve \textbf{generalized} category discovery and how to design modality-agnostic solutions without task-specific fine-tuning.

In this work, we propose OmniGCD as a method capable of performing GCD at this abstract, modality-agnostic level. Furthermore, we complete the GCD task in a zero-shot manner, where the method performs the GCD task without modality or dataset specific fine-tuning. Towards this end, we utilize pre-trained modality-specific encoders which are analogous to human sensory inputs. Similar to humans, performance is reliant on quality of the sensory inputs, and better eyes (vision encoders), such as DINOv2~\cite{dinov2} compared to DINOv1~\cite{dinov1}, achieve better results. Each of these modalities are then mapped into a unique latent space, which we refer to as a GCD latent space. Ideally, this GCD latent space should be optimal for clustering, allowing easy label assignments, thereby improving GCD performance. To achieve this, we introduce a novel GCD Transformer (GCDformer) which is trained to transform the GCD latent space into a representation better suited for clustering at test-time. A summary of our approach is shown in Figure \ref{fig:summary}. This design explicitly separates representation learning from category discovery, allowing improvements in encoder quality to directly translate into improved GCD performance.

GCDformer is trained using synthetically generated data. We find that one challenge of generating synthetic training data is ensuring sampling coverage and diversity as the GCDformer is quick to overfit to repeated data distributions. To address this, we opt to ensure that the sampling of the GCD latent space be tractable and therefore of a low-dimensionality. This requires the use of a dimension reduction method for mapping feature vectors from modality-specific encoders into a GCD latent space. Taking this modality-agnostic approach, we demonstrate that a single GCDformer model, \textbf{trained only once using synthetic data}, can perform GCD across the visual, audio, text and remote sensing modalities in a zero-shot manner, without any modality or dataset specific fine-tuning. Additionally, to our knowledge, the proposed OmniGCD is the first method to perform GCD in the audio modality.

We present zero-shot GCD results across 16 datasets spanning four modalities, showing improvement over measured baselines and setting a benchmark for future modality-agnostic GCD methods. We find that taking this approach highlights the need for strong modality-specific encoders to produce high-quality representations. As these representations form the basis on which the abstracted GCD reasoning can be performed. However, abstracting the GCD task fundamentally decouples it from simply seeking to improve the quality of the encoder. Adopting a modality-agnostic approach therefore allows performance gains to improve GCD abilities across all modalities.


\noindent
\textbf{Contributions - } We list our key contributions below: 
\begin{enumerate}
    \item We introduce zero-shot GCD, where models perform category discovery without any dataset-specific fine-tuning, while allowing modality-specific pre-trained encoders.
    \item We propose OmniGCD, a method for modality-agnostic zero-shot GCD. OmniGCD utilizes a novel GCD transformer, GCDformer, trained once on synthetic data to transform the GCD latent space into a representation better suited for clustering at test-time.
    \item We present results on 16 datasets across four unique modalities. OmniGCDs improves performance on modality-agnostic zero-shot GCD compared to baselines. Our results set a benchmark for future modality-agnostic GCD work. 
\end{enumerate}

%% file: sec/2_related_works.tex
\section{Related Work}

\subsection{Generalized Category Discovery}
Generalized Category Discovery (GCD), introduced in \cite{GCD2022}, is a human-like image classification task where models predict image clusters using a set of partially labeled images. As an extension of novel category discovery, GCD research is growing, with works addressing issues like bias mitigation~\cite{guo2022robust, liu2025debgcd, PIM23, muGCD23, simgcd}, class number estimation~\cite{zhao2023learning, liu2023open, choi2024contrastive}, label assignment and clustering~\cite{otholt2024guided, du2023fly, zheng2024prototypical}, and integration of external knowledge via LLMs and VLMs~\cite{ouldnoughi2023clip, zheng2024textual, wang2025get, yang2025consistent}. Some combine GCD with tasks like continual learning~\cite{cendra2024promptccd, wu2023metagcd, ma2024happy} and active learning~\cite{ma2024active, xie2024deep}, significantly improving state-of-the-art accuracy for visual GCD. However, these methods only apply to the visual GCD modality. Text modality GCD tasks \cite{glean, loop} have gone down their own path, solving modality-specific challenges. Likewise for the remote sensing/aerial modality~\cite{xu2024generalized, zhou2024generalized}. While solving these modality-specific challenges is undoubtedly important, this risks siloing progress to their respective modalities. As an example ConGCD~\cite{tang2025dissecting} uses visual primitives for class prediction, which cannot be directly applied to other modalities. This highlights the problem in GCD works that we address, as we seek to construct a modality-agnostic GCD method. 

\subsection{Transformers as Set Processors}
Transformers function as inherent set processors, adapting mainly through modifications to their prediction mechanisms. They can operate in auto-regressive (causal) mode, like large language models predicting the next token, or non-causal mode, like the Vision Transformer (ViT)~\cite{vit20}, which tokenizes image patches for full-attention class prediction. Their versatility, processing tokenized inputs across data types, has driven state-of-the-art results in reinforcement learning (Decision Transformer~\cite{DiT21}), graph data (Graph Transformer~\cite{graphformer19}), and chemical reactions (Molecular Transformer~\cite{molecular19}). Recent applications include abstract set processing in representation learning ~\cite{croco22, assran2023self} for tasks like depth estimation, and pose reconstruction (RelPos++~\cite{relpos2024}, ADEN~\cite{aden2024}, SparsePose~\cite{sinha2023sparsepose}, PoseDiffusion~\cite{wang2023posediffusion}), tokenizing whole images to predict camera positions. This versatility as a generic learner motivates our use of a transformer to transform the GCD latent space at test-time, without gradient updates.


%% file: sec/3_method.tex
\section{Problem Formulation}
\label{sec:problem_formulation}
Our goal is to develop a method to perform Generalized Category Discovery (GCD) at an abstract level, mimicking how humans form abstract categories as shown in \textit{category learning}~\cite{ashby2005human}. To do so, we must first reformulate the standard GCD problem in order to more accurately align with how humans perform GCD. The standard GCD formulation is as follows.

\noindent
\textbf{Standard GCD -} We follow the problem formulation provided in \cite{GCD2022}, while expanding it from the vision-modality to any data modality. GCD is a classification problem consisting of labeled \textit{known} ($D_L$) and unlabeled \textit{unknown} sets of data, ($D_U$), $D_L=\{(x_i^l, y_i^l)\} \in X\times Y_L$ and $D_U=\{(x_i^u, y_i^u)\} \in X \times Y_U$ respectively, which are jointly derived from a dataset $D$. Note that $Y_L \subset Y_U$, and labels $y_i^u$ in $D_U$ are hidden from the model. The task is to assign labels to $D_U$ from only observing $D_L$, however, the model has access to both $D_L$ and $D_U$ during training (e.g. for fine-tuning the encoder).

We highlight that the use of $D_U$ and $D_L$ for training requires that $N$ unique models be trained for $N$ datasets, and the introduction of any new data to $D_U$ or $D_L$ requires re-training the model. We wish to train a single model capable of performing GCD given any possible set of $D_U$ and $D_L$. Therefore, we introduce the novel zero-shot GCD as follows:

\noindent\textbf{Zero-shot GCD - } In the zero-shot setting we largely follow the standard formulation, however with no access to $D_L$ and $D_U$ during training. The model can only access $D_L$ at test-time to be used in a zero-shot transductive manner, to guide the model in assigning labels to $D_U$. In this formulation, \textit{zero-shot} refers to the absence of any dataset-specific fine-tuning on $D_L$ and $D_U$, while still allowing modality-specific pre-trained encoders.

\begin{figure}[t]
    \centering
    \includegraphics[width=\linewidth]{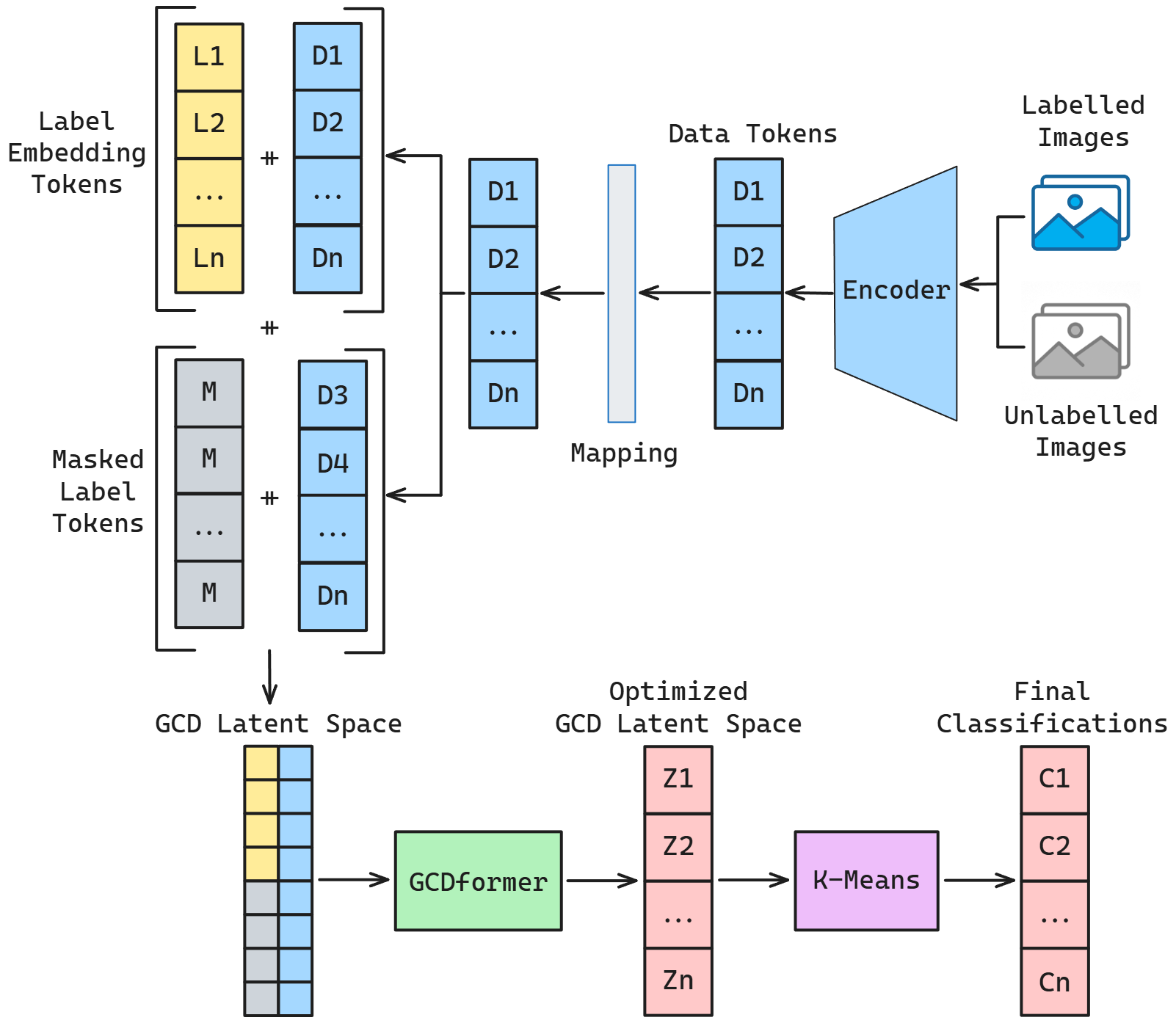}
    \caption{Overview diagram of OmniGCD. Data tokens are produced from images of the labeled and unlabeled datasets. The data tokens are then mapped to a GCD latent space and concatenated (shown as $\mdoubleplus$) with either label embedding or masked label tokens. The GCD latent space is then transformed at test-time into a representation better suited for clustering by GCDformer, before final classifications are produced by k-means clustering.}
    \label{fig:method_diagram}
\end{figure}

\section{OmniGCD}
We propose OmniGCD, a method for performing modality-agnostic zero-shot GCD. Core to the functioning of OmniGCD is the use of a novel GCD transformer, which we refer to as GCDformer. At a high-level, GCDformer is a standard Transformer, based on the GPT-2~\cite{radford_language_nodate} architecture, which takes as input a \textbf{set} of data points with associated label information, describing the labeled and unlabeled partitions of the GCD task. These data points can be derived from any source modality, making it \textbf{modality-agnostic}. We refer to the space these data points occupy as the \textbf{GCD latent space}. This latent space should be ideal for clustering, with tight clusters for each class and good separation between classes. Therefore, the objective of GCDformer is to optimizes the GCD latent spaces for these characteristics, allowing k-means clustering to easily assign labels. An overview diagram of OmniGCD is provided in Figure \ref{fig:method_diagram}. 

Developing OmniGCD requires solving various problems, with the primary challenges being \textbf{tokenization in the GCD latent space}, and \textbf{training the GCDformer}. This is elaborated on in Sections \ref{sec:arch_and_training} and \ref{sec:GCDFormer_loss} respectively. Next is the challenge of sourcing \textbf{training data}, for which we turn to \textbf{synthetic generation}, to be discussed in Section \ref{sec:synthetic_data}.

\subsection{Tokenization in the GCD Latent Space}
\label{sec:arch_and_training}
A GCD latent space refers to a latent space representing the GCD task\footnote{Each GCD task is assumed to have a unique GCD latent space. Thus, different GCD tasks do not have a shared GCD latent space.}. It contains a set of data points and their associated label information, which is either known for data in $D_L$, or unknown for data in $D_U$. OmniGCD requires that the data of the GCD latent space be tokenized for processing by GCDformer. Tokenization is performed as follows. 

For each data point $x \in D_L \cup D_U$, we construct its \textbf{data token} $T_{data}$ by computing the feature vector $T_{data} = \mathcal{E}(x) \in \mathbb{R}^{d_\mathcal{E}}$, where $\mathcal{E}$ is some modality-specific encoder and $d_\mathcal{E}$ is its output dimensionality. We employ a mapping $\mathcal{M}$ to map \textbf{data tokens} to a desired dimensionality $\mathcal{M}: \mathbb{R}^{d_\mathcal{E}} \to \mathbb{R}^{d}$. The selection of $\mathcal{M}$ and $d$ are discussed in Section \ref{sec:sample_space}. GCDformer needs to differentiate the labeled data of $D_L$ from the unlabeled data of $D_U$. Moreover, the labeled data must contain information regarding which label it belongs to. To do so, we concatenate the label information, as a \textbf{label token} $T_{label}$, directly to the \textbf{data token} of each data point $x$. The \textbf{label token} is either: (1) a \textbf{label embedding token} $T_{LE} \in \mathbb{R}^{d_l}$, for $x\in D_L$, or (2) a learnable \textbf{masked label token} $T_{LM}\in \mathbb{R}^{d_l}$, for $x \in D_U$. Where $d_l$ is the dimensionality of the \textbf{label token}, a hyperparameter of GCDformer. The \textbf{label embedding token} follows the same formulation as sinusoidal position embeddings~\cite{attention17}, however encoding label information instead of positional information as follows:

\begin{align*}
    T_{LE}(y^l, 2i) &= \sin(y^l/10000^{2i/d_{l}}) \\
    T_{LE}(y^l,2i+1) &= \cos(y^l/10000^{2i/d_{l}})
\end{align*}

\noindent
where $y^l$ is the integer label and $i$ is the dimension. Due to the concatenation of $T_{data}$ and $T_{label}$, the resulting internal dimensionality of the GCDformer is $d_{model}=d_d+d_l$. We do not encode any positional information, as the position of each token holds no value in the GCD task. This is due to each input being a \textbf{set} as opposed to a \textbf{sequence}. With this, the GCD task as contained in the GCD latent space can be converted into a set of tokens for processing by the GCDformer. The training of GCDformer is described in the following section.

\subsection{Training the GCDformer}
\label{sec:GCDFormer_loss}
GCDformer performs a learned transformation of the GCD latent space by projecting features guided by information encoded in the \textbf{data} and \textbf{label tokens}. The resulting latent space is better suited for clustering, allowing k-means to more accurately assign cluster labels. Figure \ref{fig:summary} shows one example of this, with more qualitative examples of the improvements in the GCD latent spaces provided in the supplementary material. 

Here we describe the models training objective given the tokenized data of a GCD latent space (Section \ref{sec:arch_and_training}). As the GCD latent space now represents a \textbf{set} of tokens, GCDformer therefore employs non-causal self-attention, allowing all tokens to attend to each other simultaneously.

\noindent
\textbf{Training Loss - } Once the set has be processed by the GCDformer, we consider each token at the output to be a data point $z$, and apply contrastive loss as follows. Given a batch of $N$ samples, where each sample $i$ contains $n_i$ points with embeddings $z_j^i \in \mathbb{R}^d$ and corresponding labels $y_j^i$, the loss for each sample $i$ is defined as:

\begin{align*}
    L_i = \frac{1}{|P_i|} \sum_{(j,k) \in P_i} l(z_j^i, z_k^i, y_j^i, y_k^i)
\end{align*}

\noindent
where $P_i$ is the set of all unique pairs in sample $i$ $(j < k)$, and the pair loss function $l$ is defined as:

\begin{align*}
    l(z_j^i, z_k^i, y_j^i, y_k^i) = \begin{cases} 
      \|z_j^i - z_k^i\|_2^2 & \text{if } y_j^i = y_k^i \\
      \max(0, m - \|z_j^i - z_k^i\|_2)^2 & \text{if } y_j^i \neq y_k^i 
   \end{cases}
\end{align*}

\noindent
where $m$ is the margin parameter, enforcing desired separability. The final loss across the batch is:

\begin{align*}
    L = \frac{1}{N} \sum_{i=1}^N L_i
\end{align*}

\noindent This objective directly enforces intra-class compactness and inter-class separation, aligning with the clustering objective of GCD.



\subsection{Synthetic Training Data}
\label{sec:synthetic_data}


To ensure OmniGCD remains modality-agnostic and unbiased towards any specific modality, we utilize synthetic data for training GCDformer. We design synthetic data to satisfy two key properties: (1) sufficient coverage of the GCD latent space, and (2) alignment with real data distributions. These properties are sufficient for learning a transformation that improves clustering in the latent space. 
Addressing (1), we opt to use a low-dimensional GCD latent space, which we elaborate on in Section~\ref{sec:sample_space}. This requires the use of a dimension reduction method as the mapping $\mathcal{M}$. An ideal method produces a GCD latent space with low local information loss, high cluster separation, high cluster spread, and low cluster overlap, while maintaining alignment with real data.
Importantly, our goal is not to model the full real-world data distribution, but to capture the structural properties of the GCD latent space required for clustering (e.g., separability and coverage). As shown in Section~\ref{sec:dim_reduction}, these properties translate to improved clustering performance across modalities, supporting the sufficiency of the synthetic data.

\subsubsection{Tractable Sampling Space}
\label{sec:sample_space}
Our goal is to enable tractable sampling coverage for generating the synthetic data. This requires data points to cover the sampling space $S = [-1,1]^d$, where $d$ is the data dimensionality. Due to the curse of dimensionality~\cite{dimcurse}, covering $S$ becomes intractable as $d$ increases. This can be shown by considering the sampling of a uniform grid in a $d$-dimensional unit hypercube $[0,1]^d$ with spacing $\epsilon$ between adjacent points, where $\epsilon>0$. In this case, the number of points ($N$) needed per dimension is approximately $\frac{1}{\epsilon}+1$, which simplifies to $N\approx(\frac{1}{\epsilon})^d$ for small values of $\epsilon$. This shows that $N$ grows exponentially with $d$. To minimize this issue and ensure effective coverage, we should select a low-dimensional sampling space. Importantly, this reduced space need only preserve cluster structure rather than full feature fidelity.


\subsubsection{Dimensionality Reduction Method}
\label{sec:dim_reduction}
To utilize a low-dimensional GCD latent space for OmniGCD, we need a dimensionality reduction method for use as mapping $\mathcal{M}$, defined in Section 4.1. We prioritize non-parametric methods to avoid introducing modality- or dataset-specific parameters, which would contradict our modality-agnostic zero-shot objective. An ideal dimension reduction method should: (1) ensure high similarity between simulated and real data distributions, and (2) produce a good GCD latent space, with low local information loss, high cluster separation, high cluster spread and low cluster overlap. We examine Principal Component Analysis (PCA)~\cite{pca}, Uniform Manifold Approximation and Projection (UMAP)~\cite{umap} and t-distributed Stochastic Neighbor Embedding (t-SNE)~\cite{tsne}.

PCA, a linear dimensionality reduction method, assumes data lies on a linear subspace, making it unsuitable for capturing the inherently nonlinear relationships in a neural network's feature space when reduced to very low dimensions. This linearity constraint often fails to preserve the intrinsic geometry of the data~\cite{tsne}. Additionally, PCA's global approach does not prioritize local neighborhood preservation, resulting in clustering artifacts in low-dimensional embeddings. In contrast, UMAP and t-SNE are nonlinear methods better suited to capture complex data structures. Though UMAP and t-SNE are non-parametric, they suit our use case, processing all high-dimensional feature vectors in one pass without out-of-sample mapping. T-SNE preserves local structures and mitigates the crowding problem using a heavy-tailed Student t-distribution, ensuring low local information loss, high cluster separation, and spread, but may distort global distribution due to altered inter-cluster distances. UMAP better preserves global structure, maintaining cluster topology, positions, and orientations more effectively than t-SNE, aligning distribution similarity with balanced local and global preservation for reduced overlap and enhanced separation. We present results directly assessing PCA, UMAP and t-SNE for producing a good GCD latent space in Section \ref{sec:dim_reduction_results}.

%% file: sec/4_results.tex
\section{Experiments}
Here we first present results for selecting the dimension reduction method, before presenting OmniGCD results. OmniGCD utilizes a \textbf{single GCDformer trained only once} for GCD latent space optimization on all 16 datasets. We describe our training setup, the datasets chosen for evaluation, the encoders used for each modality and specifics of the zero-shot GCD evaluation process. Following this setup we then present and discuss our results. 

\begin{table}[t]
\centering
\resizebox{\linewidth}{!}{%
\begin{tabular}{lccccc}
\toprule
& Vision & Text & Audio & RS & Average \\
\midrule
PCA & 1.70 ± 0.69 & \textbf{1.19} ± 0.08 & 1.63 ± 0.72 & 1.51 ± 0.58 & 1.58 ± 0.63 \\
UMAP & 1.82 ± 0.71 & 1.61 ± 0.38 & 1.68 ± 0.89 & 2.29 ± 1.02 & 1.86 ± 0.76 \\
t-SNE & \textbf{1.35} ± 0.17 & 1.52 ± 0.14 & \textbf{1.34} ± 0.13 & \textbf{1.42} ± 0.3 & \textbf{1.41} ± 0.19 \\
\bottomrule
\end{tabular}%
}
\caption{Kullback-Leibler (KL) divergence of synthetic and real data using PCA~\cite{pca}, UMAP~\cite{umap} and t-SNE~\cite{tsne}. We report the average and standard deviation for each modality with lower values indicating better synthetic-to-real alignment. We \textbf{bold} the best results for each comparison. RS: Remote Sensing.}
\label{tab:KS-KL}

\end{table}

\begin{table}[t]
\resizebox{\linewidth}{!}{%
\begin{tabular}{@{\extracolsep{4pt}}lccccccccc@{}}
\toprule
 & \multicolumn{3}{c}{Cluster Separation ($\uparrow$)} & \multicolumn{3}{c}{Cluster Spread ($\uparrow$)} & \multicolumn{3}{c}{Cluster Overlap ($\downarrow$)} \\
 \cline{2-4} \cline{5-7} \cline{8-10}
\rule{0pt}{3ex} & PCA & t-SNE & UMAP & PCA & t-SNE & UMAP & PCA & t-SNE & UMAP \\
 \midrule
Vision & 0.62 & \underline{0.74} & \textbf{0.75} & \textbf{0.24} & \underline{0.22} & 0.19 & 12.52 & \textbf{10.33} & \underline{10.44} \\
Text & 0.65 & \textbf{0.85} & \underline{0.81} & \textbf{0.22} & \underline{0.11} & 0.09 & 4.04 & \textbf{1.06} & \underline{1.07} \\
Audio & \underline{0.43} &\textbf{0.53} & 0.27 & 0.38 & \underline{0.45} & \textbf{0.46} & \textbf{6.15} & \underline{12.84} & 29.32 \\
RS & 0.57 & \textbf{0.82} & \underline{0.77} & \textbf{0.29} & \underline{0.24} & 0.17 & 3.99 & \underline{1.29} & \textbf{1.27} \\
\midrule
Average & 0.57 & \textbf{0.74} & \underline{0.65} & \textbf{0.28} & \underline{0.26} & 0.23 & \underline{6.68} & \textbf{6.38} & 10.53 \\
\bottomrule
\end{tabular}%
}
\caption{Comparison of PCA, t-SNE, and UMAP across modalities for Cluster Distance (inter-cluster separation), Cluster Spread (intra-cluster compactness), and Cluster Overlap (mean class overlap ratio). Higher values are better for Cluster Distance and Spread, while lower values are better for Cluster Overlap. We \textbf{bold} the best results and \underline{underline} the second best results for each comparison. RS: Remote Sensing.} 
\label{tab:cluster}
\end{table}

\subsection{Dimension Reduction Results}
\label{sec:dim_reduction_results}
We evaluate three dimensionality reduction methods—Principal Component Analysis (PCA)~\cite{pca}, Uniform Manifold Approximation and Projection (UMAP)~\cite{umap} and t-distributed Stochastic Neighbor Embedding (t-SNE)~\cite{tsne}—based on (1) their ability to align synthetic and real GCD latent spaces and (2) produce latent spaces suitable for clustering. To measure synthetic-to-real alignment, we compute the Kullback-Leibler (KL) divergence between synthetic and real GCD latent spaces. Results in Table \ref{tab:KS-KL}, averaged across 1,000 synthetic latent spaces per modality, show t-SNE achieves the lowest KL divergence (1.41).

To evaluate clustering suitability, we measure cluster distance (inter-cluster separation), cluster spread (intra-cluster compactness), and cluster overlap (mean class overlap ratio). An ideal method exhibits high separation and spread with low overlap. Table \ref{tab:cluster} shows that t-SNE achieves the best balance across these metrics. Crucially, these properties translate to improved GCD performance (Section \ref{sec:ab_dim_reduction}), supporting that alignment and clustering structure in the synthetic latent space are sufficient for effective GCD. Therefore, we select t-SNE as $\mathcal{M}$ for all following results.

\subsection{Setup}

\textbf{Model and Training - } GCDformer consists of 6 attention block layers with 4 self-attention heads each, an internal embedding dimension ($d_{model}$) of 256, of which the \textbf{label token} accounts for 32 dimensions and \textbf{data token} accounts for the remaining 224 dimensions. To ensure tractability (Section \ref{sec:sample_space}) during data generation, we opt for generating 2D data. We provide ablation results using higher input and output dimensions in the supplementary material. The model is trained for 8,000 epochs with a constant learning rate of 0.0001 using the AdamW optimizer~\cite{adamoptim}, and a batch size of 16. We generate up to 200 clusters ($N_{cluster}$) during training with a maximum set size of up to 3,000 points. 

\noindent
\textbf{Training Data Generation - } The synthetic GCD latent space generation process consists of two stages: (1) generation of data and (2) masking of labels. In the generation stage, we generate a random number of clusters, up to $N_{cluster}$. Cluster centers and points are sampled from Gaussian-based (normal, Laplace, or von Mises) or uniform distributions, with each cluster being randomly assigned a unique integer label between 1 to 1000 (0 is reserved for masked points). We randomly mask a percentage of points within each cluster and fully mask a random percentage of clusters. This produces diverse training examples with varying set sizes, cluster counts, and masking.

\noindent\textbf{Datasets - } We conduct tests on 16 total datasets across the vision, text, audio and remote sensing modalities. For vision, we test using the following classification datasets: CIFAR-10 and CIFAR-100~\cite{cifar}, ImageNet-100~\cite{imagenet}, CUB-200~\cite{cub200}, Stanford Cars~\cite{stanford_cars}, FGVC-Aircraft~\cite{aircraft} and Herbarium-19~\cite{herb19}. For the text modality, we use BANKING~\cite{bankingdset}, StackOverflow~\cite{stackoverflow} and CLINIC~\cite{clinic} which all contain labeled text data for tasks such as sentiment analysis, question classification, and intent recognition, respectively. In the audio modality, we use VocalSet~\cite{vocalset} and UrbanSound~\cite{urbansound}, which contain recordings of vocal techniques and urban environmental sounds for classification tasks. Lastly, for the remote sensing modality we use EuroSAT~\cite{eurosat}, So2SAT~\cite{so2sat}, RESISC45~\cite{resic45} and UC Merced~\cite{ucmerced}, which contain satellite imagery for land use and land cover classification tasks. While similar to the vision modality, the remote sensing modality utilities two multi-spectral datasets (EuroSAT and So2SAT), and uses a hyperspectral encoder model. The number of labeled and unlabeled classes for each dataset are shown in Table \ref{tab:num_classes}.

\noindent\textbf{Encoders - } OmniGCD relies on the use of pre-trained encoder for each modality. For the vision modality, we conduct tests using both the DINOv1~\cite{dinov1} and DINOv2~\cite{dinov2} vision encoders, as is done in prior GCD works. For text, audio and remote sensing we use the E5-Large-v2~\cite{e5text} text encoder, MERT-95M~\cite{mertaudio} audio encoder and DOFA-Base~\cite{dofa} hyperspectral encoder. More results for additional encoders are provided in the supplementary material.

\begin{table}[t]
\resizebox{\linewidth}{!}{%
\begin{tabular}{lcccccccc}
\toprule
 & CIFAR-10 & CIFAR-100 & ImageNet-100 & CUB-200 & SCars & Aircraft & Herb-19 & BANKING \\
 \midrule
$|{Y_l}|$ & 5 & 80 & 50 & 100 & 98 & 51 & 341 & 38 \\
$|Y_u|$ & 10 & 100 & 100 & 200 & 196 & 102 & 693 & 77 \\
\midrule
 & StackOverflow & CLINIC & VocalSet & UrbanSound & EuroSAT & So2SAT & RESISC45 & UC Merced \\
 \midrule
$|Y_l|$ & 10 & 75 & 9 & 5 & 5 & 7 & 15 & 9 \\
$|Y_u|$ & 20 & 150 & 18 & 10 & 10 & 15 & 45 & 21 \\
\bottomrule
\end{tabular}%
}
\caption{Number of classes in the labeled and unlabeled sets ($|Y_l|$, $|Y_u|$) for each dataset.}
\label{tab:num_classes}
\end{table}

\begin{table*}[t]
\resizebox{\linewidth}{!}{%
\begin{tabular}{@{\extracolsep{4pt}}llccccccccccccccccccccc}
\toprule
 &  & \multicolumn{3}{c}{\faEye[regular] \space CIFAR-10~\cite{cifar}} & \multicolumn{3}{c}{\faEye[regular] \space CIFAR-100~\cite{cifar}} & \multicolumn{3}{c}{\faEye[regular] \space ImageNet-100~\cite{imagenet}} & \multicolumn{3}{c}{\faEye[regular] \space CUB-200~\cite{cub200}} & \multicolumn{3}{c}{\faEye[regular] \space Stanford Cars~\cite{stanford_cars}} & \multicolumn{3}{c}{\faEye[regular] \space Aircraft~\cite{aircraft}} & \multicolumn{3}{c}{\faEye[regular] \space Herbarium-19~\cite{herb19}} \\ \cline{3-5}\cline{6-8}\cline{9-11}\cline{12-14}\cline{15-17}\cline{18-20}\cline{21-23}
\rule{0pt}{11pt} &  & All & Old & New & All & Old & New & All & Old & New & All & Old & New & All & Old & New & All & Old & New & All & Old & New \\ \midrule

\multirow{3}{*}{\rule{0pt}{2ex} \rotatebox{90}{\scriptsize DINOv1~\cite{dinov1}}} & K-means & 77.8 & \underline{88.1} & 60.3 & 52.5 & 68.9 & 66.3 & \underline{73.5} & 87.6 & 69.9 & 35.8 & 54.7 & \textbf{39.8} & 10.7 & 14.5 & 15.9 & 14.8 & 10.9 & \underline{10.8} & 24.9 & 42.9 & \underline{22.5} \\

 & GCD (w/o FT) & \underline{81.0} & 43.9 & \underline{88.3} & \textbf{61.9} & \underline{86.5} & \underline{70.0} & 67.5 & \textbf{93.0} & \underline{72.0} & \underline{39.4} & \textbf{75.0} & 8.47 & \textbf{12.8} & \underline{17.4} & \textbf{33.3} & \underline{16.4} & \underline{13.4} & 7.6 & \underline{29.2} & \textbf{71.4} & 0.0 \\

 
 & OmniGCD & \textbf{90.7} & \textbf{97.0} & \textbf{91.6} & \underline{60.0} & \textbf{89.9} & \textbf{74.6} & \textbf{81.1} & \underline{91.0} & \textbf{74.8} & \textbf{44.5} & \underline{66.0} & \underline{33.9} & \underline{12.6} & \textbf{19.6} & \underline{16.9} & \textbf{18.9} & \textbf{14.3} & \textbf{13.9} & \textbf{30.8} & \underline{48.6} & \textbf{50.0} \\
 \midrule
 
\multirow{3}{*}{\rule{0pt}{3ex} \rotatebox{90}{\scriptsize DINOv2~\cite{dinov2}}} & K-means & 82.1 & 62.6 & 95.0 & 69.1 & \underline{44.2} & \underline{51.3} & \underline{79.9} & 87.7 & 78.4 & \underline{70.3} & \underline{95.8} & \underline{75.4} & 27.3 & \underline{16.8} & 28.7 & 19.6 & 12.7 & \textbf{22.5} & 29.0 & \underline{40.8} & 23.2 \\

 & GCD (w/o FT) & \underline{84.9} & \underline{86.4} & \textbf{96.4} & \underline{71.4} & 39.0 & 32.5 & 67.6 & \underline{92.0} & \underline{84.0} & 66.3 & \textbf{100.0} & 53.6 & \underline{31.8} & \underline{19.5} & \underline{29.1} & 19.5 & \underline{13.6} & 17.9 & \textbf{34.9} & 30.8 & \textbf{45.2} \\
 
 & OmniGCD & \textbf{96.9} & \textbf{96.9} & \underline{95.6} & \textbf{78.1} & \textbf{47.9} & \textbf{56.8} & \textbf{88.7} & \textbf{94.0} & \textbf{90.0} & \textbf{79.8} & \textbf{100.0} & \textbf{96.4} & \textbf{33.4} & \textbf{24.2} & \textbf{43.1} & \textbf{21.2} & \textbf{17.0} & 11.6 & \underline{34.8} & \textbf{60.0} & \underline{25.8} \\
 \midrule
 &  & \multicolumn{3}{c}{\faFile*[regular] BANKING~\cite{bankingdset}} & \multicolumn{3}{c}{\faFile*[regular] StackOverflow~\cite{stackoverflow}} & \multicolumn{3}{c}{\faFile*[regular] CLINIC~\cite{clinic}} & \multicolumn{3}{c}{\faMicrophone \space VocalSet~\cite{vocalset}} & \multicolumn{3}{c}{\faMicrophone \space UrbanSound~\cite{urbansound}} & \multicolumn{3}{c}{\faSatellite \space EuroSAT~\cite{eurosat}} & \multicolumn{3}{c}{\faSatellite \space So2SAT~\cite{so2sat}} \\ \cline{3-5}\cline{6-8}\cline{9-11}\cline{12-14}\cline{15-17}\cline{18-20}\cline{21-23}
\rule{0pt}{11pt} & & All & Old & New & All & Old & New & All & Old & New & All & Old & New & All & Old & New & All & Old & New & All & Old & New \\ \midrule

 & K-means & \underline{57.7} & 64.4 & \underline{43.4} & \underline{72.9} & \underline{68.3} & \underline{45.2} & \underline{69.5} & \underline{85.0} & \underline{81.7} & 22.1 & 15.0 & 25.0 & 39.3 & 37.6 & \underline{37.7} & \underline{53.2} & \underline{38.4} & 62.4 & 27.3 & 32.5 & \underline{65.1}\\
 
 & GCD (w/o FT) & 57.3 & \textbf{96.3} & \textbf{83.8} & 60.1 & 47.0 & 0.0 & 64.3 & 56.7 & \textbf{90.0} & \underline{25.5} & \textbf{17.7} & \underline{26.0} & \underline{43.0} & \underline{49.7} & 9.26 & 52.8 & 38.2 & \underline{62.6} & \underline{27.5} & \underline{34.4} & \textbf{66.2} \\
 
 & OmniGCD & \textbf{66.4} & \underline{77.1} & 40.8 & \textbf{86.8} & \textbf{78.0} & \textbf{73.8} & \textbf{82.1} & \textbf{96.0} & \textbf{90.0} & \textbf{25.6} & \underline{17.5} & \textbf{49.2} & \textbf{45.5} & \textbf{56.7} & \textbf{63.9} & \textbf{68.3} & \textbf{75.5} & \textbf{69.5} & \textbf{31.7} & \textbf{38.4} & 58.4  \\
 
 \midrule
 &  & \multicolumn{3}{c}{\faSatellite \space RESISC45~\cite{resic45}} & \multicolumn{3}{c}{\faSatellite \space UC Merced~\cite{ucmerced}} \\ \cline{3-5}\cline{6-8}
\rule{0pt}{11pt}  & & All & Old & New & All & Old & New  \\ \cline{3-8}
\rule{0pt}{11pt} & K-means & \underline{42.8} & \underline{37.1} & \underline{60.8} & \underline{63.6} & 50.6 & \underline{70.8}  \\
 & GCD (w/o FT) & 42.7 & 35.5 & 55.6 & 60.5 & \underline{55.9} & 51.0  \\
 & OmniGCD & \textbf{58.5} & \textbf{73.0} & \textbf{74.4} & \textbf{75.8} & \textbf{67.7} & \textbf{95.7} \\ \bottomrule
\end{tabular}%
}

\caption{Zero-shot GCD results across all modalities. Vision (\faEye[regular]) uses DINOv1 (ViT-B/16)~\cite{dinov1} and DINOv2 (ViT-B/14)~\cite{dinov2} vision encoders. Text (\faFile*[regular]) uses the E5-Large~\cite{e5text} text encoder. Audio (\faMicrophone) uses the MERT-95M~\cite{mertaudio} audio encoder. Remote sensing (\faSatellite) uses the DOFA-Base~\cite{dofa} hyperspectral encoder. We present results using the GCD evaluation metrics comparing K-means~\cite{kmeans1982} and GCD~\cite{GCD2022} without fine-tuning (GCD (w/o FT)). Each result is the average of five independent runs, with standard deviation reported in the supplementary material. OmniGCD results are gathered using the same GCDformer model weights (only trained once). OmniGCD improves average \textit{All} accuracy across datasets within each modality by $+6.2$, $+17.9$, $+1.5$ and $+12.7$ points for vision, text, audio and remote sensing modalities respectively. We \textbf{bold} the best results and \underline{underline} the second best results for each comparison.}
\label{tab:zero-shot_vision}
\end{table*}

\noindent\textbf{Zero-shot GCD Evaluation - } During test-time, the entire input for OmniGCD contains both the known and unknown subsets of data. For the zero-shot GCD setting, the known subset is constructed using data from the training split which belongs to the known classes and the unknown subset is the test set. We report standard GCD metrics, where \textit{All} is the overall classification accuracy on all classes; \textit{Old} is the classification accuracy on known classes; and \textit{New} is the classification accuracy of unknown classes. All reported results are the average values for five independent runs, with standard deviation values reported in the supplementary material. Additionally, we provide a more detailed analysis of OmniGCD results in the supplementary material. For baseline methods, we compare our results against (1) k-means clustering of the modality-specific feature representations (\textbf{K-means}), and (2) the original GCD method~\cite{GCD2022} without fine-tuning (\textbf{GCD (w/o FT)}). Since prior GCD methods rely on dataset-specific fine-tuning, removing this step reduces them to clustering on frozen encoder features (e.g., semi-supervised k-means), making them appropriate baselines under the zero-shot constraint. We provide comparisons to state-of-the-art GCD methods in Section \ref{sec:SOTA}.

\subsection{Zero-shot GCD Results}
\label{sec:zs_results}
In Table \ref{tab:zero-shot_vision} we present zero-shot GCD results across 16 tested datasets using five different pre-trained encoders belonging to four unique modalities. 

\noindent
\textbf{Vision - }Firstly, using the DINOv1~\cite{dinov1} encoder, OmniGCD achieves the best \textit{All} class accuracy on 5 of the 7 tested datasets. GCD (w/o FT) only slightly outperforms it on CIFAR-100 and Stanford Cars. Of the instances belonging to \textit{Old} classes, OmniGCD achieves the best results on 4 of the 7 datasets, while for the classes belonging to the \textit{New} we achieve the best results on  5 of the 7 datasets. OmniGCD shows improved performance when utilizing the stronger DINOv2~\cite{dinov2} vision encoder, achieving the best results on 6 of the 7 datasets for \textit{All}, 7 out of 7 for \textit{Old} and 4 out of 7 for \textit{New}. Interestingly, on the FGVC-Aircraft dataset k-means achieves the best performance on the \textit{New} classes. The relative improvement in performance between DINOv1 and DINOv2 shows that OmniGCD performance improves with the quality of the encoder utilized. 

\noindent
\textbf{Text - } Next, we analyze the text modality datasets, using the same GCDformer used for the vision modality. OmniGCD performs significantly better for \textit{All} classes on every text dataset. GCD (w/o FT) only outperforming OmniGCD on the \textit{Old} and \textit{New} classes on BANKING, and matches \textit{New} class accurcy for CLINIC.

\noindent
\textbf{Audio - } In the audio modality, OmniGCD has the highest accuracy for every metric except on the \textit{Old} classes of VocalSet, although it is only a 0.2 points behind GCD (w/o FT). OmniGCD's main improvement is on discovering \textit{New} classes, where we see a +23.2 point improvement on VocalSet, and a +26.2 point improvement on UrbanSound. This highlights the ability of OmniGCD to discover novel classes and demonstrates that fine-tuning is not the only avenue for achieving strong GCD performance.

\noindent
\textbf{Remote Sensing - } For remote sensing, OmniGCD is the best method across every dataset, again only being beaten once by GCD (w/o FT) on the \textit{New} classes of So2SAT. However, in this modality, GCD (w/o FT) generally struggles to improve performance over k-means. 
 

%% file: sec/5_ablations.tex
\section{Discussion and Ablations}
In this section, we provide additional results and a discussion regarding (1) if we still require fine-tuning for GCD, and (2) the gap between our proposed OmniGCD method for zero-shot GCD and the current state-of-the-art methods for standard GCD. We also present ablation results regarding the selection of t-SNE for dimension reduction.

\subsection{Do we still require fine-tuning for GCD?}
\label{sec:do_we_need_ft}
OmniGCD is designed to decouple representation learning from category discovery. As with all GCD methods, performance depends on the quality of the underlying encoder; however, unlike prior work, OmniGCD operates on frozen representations without requiring dataset-specific fine-tuning. This separation allows improvements in encoder quality to directly translate to improved GCD performance without retraining the discovery module.

This dependency can be observed when using weaker encoders such as DINOv1 on fine-grained datasets (e.g., Stanford Cars and FGVC Aircraft), where performance degrades due to poor feature representations. This is further supported by comparing OmniGCD to k-means operating on the same feature space (Table~4), where consistent improvements indicate that gains are not solely due to the encoder.

In such cases, fine-tuning may still be beneficial to improve feature quality. In Table~5, we present results where the encoder is fine-tuned following the GCD~\cite{GCD2022} procedure. These results show consistent improvements, particularly on fine-grained datasets. However, additional results in the supplementary material indicate that fine-tuning does not always lead to consistent gains. This suggests that while encoder quality remains important, OmniGCD can effectively leverage strong representations without requiring dataset-specific adaptation.

\subsection{Gap between OmniGCD and State-of-the-art}
\label{sec:SOTA}
As our main problem setting is zero-shot GCD, direct comparison to state-of-the-art (SOTA) methods is not possible, as they rely on dataset-specific fine-tuning and violate the zero-shot constraint described in Section \ref{sec:problem_formulation}.
However, for completeness, we compare and discuss OmniGCDs zero-shot GCD results in relation to the current SOTA in the vision and text modalities (Tables \ref{tab:sota_vision} and \ref{tab:sota_text} respectively). We provide additional results for remaining vision datasets and the remote sensing modality in the supplementary material. For vision, OmniGCD with DINOv2 achieves SOTA performance on the \textit{Old} and \textit{New} classes for CUB-200, however is -$10.9$ points in \textit{All} accuracy. For a more competitive comparison to the SOTA we include results using the recently released DINOv3~\cite{dinov3} vision encoder. We include additional DINOv3 results in the supplementary material. Using this stronger encoder, OmniGCD achieves SOTA performance on discovering the \textit{New} classes in Stanford Cars. For the text modality comparison, in Table \ref{tab:sota_text}, OmniGCD achieves SOTA results for overall accuracy (equivalent to \textit{All}) and the Adjusted Rand Index (ARI) for the StackOverflow dataset. ARI and Normalized Mutual Information (NMI) are reported here as these metrics are used for text modality GCD reporting~\cite{glean, loop}. 

The areas where OmniGCD achieves SOTA performance, despite being a zero-shot method, highlights the potential of our proposed modality-agnostic zero-shot GCD approach. Additionally, the modality-agnostic approach decouples the encoder from the GCD task, allowing improvements to be explored independently. This is beneficial as improvements in modality-agnostic performance will propagate improved GCD abilities to all modalities, whereas improvement in the modality specific GCD methods may only benefit their target modality. One avenue for future work is end-to-end training of OmniGCD, not just GCDFormer.

\begin{table}[t]
\resizebox{\linewidth}{!}{%
\begin{tabular}{@{\extracolsep{4pt}}lcccccccccccc}
\toprule
 & \multicolumn{3}{c}{\faEye[regular]\space CUB-200} & \multicolumn{3}{c}{\faEye[regular]\space Stanford Cars} & \multicolumn{3}{c}{\faEye[regular]\space FGVC Aircraft}  \\ \cline{2-4}\cline{5-7}\cline{8-10}
\rule{0pt}{2ex} & All & Old & New & All & Old & New & All & Old & New  \\ \midrule
 GCD~\cite{GCD2022} & 51.3 & 56.6 & 48.7 & 39.0 & 57.6 & 29.9 & 45.0 & 41.1 & 46.9 \\
 OmniGCD (w/o FT) & 44.5 & 66.0 & 33.9 & 12.6 & 19.6 & 16.9 & 18.9 & 14.3 & 13.9 \\
\midrule
 OmniGCD (w FT) & \textbf{59.8} & \textbf{71.0} & \textbf{90.6} & \textbf{49.1} & \textbf{63.7} & \textbf{48.1} & \textbf{47.5} & \textbf{71.21} & \textbf{59.4} \\ \bottomrule
\end{tabular}%
}
\caption{Results on the standard GCD setting compared to the original GCD method which fine-tunes its encoder~\cite{GCD2022}. We fine-tune the vision encoder using the same method as the original GCD method~\cite{GCD2022}. We \textbf{bold} the best results for each comparison.}
\label{tab:vis_ft_dinov2}
\end{table}

\begin{table}[t]
\resizebox{\linewidth}{!}{%
\begin{tabular}{@{\extracolsep{4pt}}lcccccccccccc}
\toprule
 & &  \multicolumn{3}{c}{\faEye[regular]\space CUB-200} & \multicolumn{3}{c}{\faEye[regular]\space Stanford Cars} & \multicolumn{3}{c}{\faEye[regular]\space FGVC Aircraft} \\ \cline{3-5}\cline{6-8}\cline{9-11}
\rule{0pt}{2ex} & Encoder & All & Old & New & All & Old & New & All & Old & New \\ \midrule
 ConGCD~\cite{tang2025dissecting}  & DINOv2 & 81.7 & 80.4 & 82.4 & 57.5 & 77.5 & 47.9 & 62.5 & 70.2 & 58.7 \\ 
 SelEx~\cite{rastegar2024selex}  & DINOv2 &87.4 & 85.1 & 88.5 & 82.2 & \textbf{93.7} & 76.7 & 79.8 & \textbf{82.3} & 78.6 \\
 HypGCD~\cite{liu2025hyperbolic}  & DINOv2 &\textbf{90.7} & 85.3 & 93.4 & \textbf{83.8} & 93.3 & 79.2 & \textbf{83.4} & 82.0 & \textbf{84.1}  \\ 
\midrule
 OmniGCD & DINOv2  & 79.8 & \textbf{100.0} & \textbf{96.4} & 33.4 & 24.2 & 43.1 & 21.2 & 17.0 & 11.6 \\
 OmniGCD & DINOv3 & 79.1 & 79.3 & 86.7 & 67.3 & 81.0 & \textbf{87.3} & 44.0 & 32.3 & 59.7 \\ \bottomrule
\end{tabular}%
}
\caption{Comparison of OmniGCDs zero-shot performance using the DINOv2 (ViT-B/14)~\cite{dinov2} and DINOv3 (ViT-B/16)~\cite{dinov3} encoders to the current state-of-the-art works in the vision modality. These methods perform GCD under the standard setting and allow fine-tuning, whereas OmniGCD does not. We \textbf{bold} the best results for each comparison.}
\label{tab:sota_vision}
\end{table}

\begin{table}[t]
\resizebox{\linewidth}{!}{%
\begin{tabular}{lccccccccc}
\toprule
 & \multicolumn{3}{c}{\faFile*[regular] BANKING} & \multicolumn{3}{c}{\faFile*[regular] StackOverflow} & \multicolumn{3}{c}{\faFile*[regular] CLINIC} \\
  \cmidrule(lr){2-4} \cmidrule(lr){5-7} \cmidrule(lr){8-10}
 & ACC & ARI & NMI & ACC & ARI & NMI & ACC & ARI & NMI \\
\midrule
GCD~\cite{GCD2022} &  74.4 & 63.8 & 84.8 & 85.6 & 72.2 & 80.1 & 86.5 & 81.2 & 94.6 \\
SimGCD~\cite{simgcd} & 74.4 & 64.2 & 85.1 & 82.0 & 70.7 & 80.4 & 87.2 & 81.7 & 94.8 \\
Loop~\cite{loop} & \underline{75.1} & \underline{65.7} & \underline{85.4} & 85.9 & 72.5 & 80.6 & \underline{91.0} & \underline{85.2} & \underline{95.6} \\
Glean~\cite{glean} & \textbf{80.3} & \textbf{70.4} & \textbf{87.7} & \underline{89.4} & \underline{78.9} & \textbf{85.0} & \textbf{94.5} & \textbf{90.8} & \textbf{97.1} \\
\midrule
OmniGCD & 67.8 & 60.2 & 81.8 & \textbf{90.1} & \textbf{81.6} & \underline{83.0} & 87.3 & 84.4 & 94.1 \\ 
\bottomrule
\end{tabular}%
}
\caption{Comparing OmniGCD Zero-shot GCD results against current state-of-the-art methods under the standard GCD setting in the text modality. Again, for our OmniGCD, no fine-tuning of the E5-Large-V2~\cite{e5text} text encoder was conducted. We \textbf{bold} the best results and \underline{underline} the second best results for each comparison.\vspace{-15pt}}
\label{tab:sota_text}
\end{table}

\subsection{Dimension Reduction Options}
\label{sec:ab_dim_reduction}
Lastly, we present ablation results related to the discussion in Section \ref{sec:dim_reduction}, where we consider t-SNE~\cite{tsne}, PCA~\cite{pca} and UMAP\cite{umap} for selection as dimension reduction methods. Our choice of t-SNE is confirmed by our results in Table \ref{tab:dim_reduction} where we compare the performance of OmniGCD on using a subset of datasets across all modalities using the various dimensionality reduction methods. We present additional results in the supplementary material. Using t-SNE results in better performance over the other methods on 14 of the 18 presented results, with UMAP being the second best choice.
\begin{table}[t]
\resizebox{\linewidth}{!}{%
\begin{tabular}{@{\extracolsep{4pt}}lccccccccc}
\toprule
 & \multicolumn{3}{c}{\faEye[regular] \space ImageNet-100} & \multicolumn{3}{c}{\faEye[regular] \space CUB-200} & \multicolumn{3}{c}{\faFile*[regular] StackOverflow} \\ \cline{2-4}\cline{5-7}\cline{8-10}
\rule{0pt}{2.5ex} & All & Old & New & All & Old & New & All & Old & New \\ 
 \midrule
 PCA~\cite{pca} & 16.6 & 9.4 & 4.8 & 10.6 & 11.3 & 20.3 & 35.3 & 34.8 & 43.2 \\
UMAP~\cite{umap} & 77.2 & 89.0 & \textbf{77.0} & 41.3 & 60.0 & 25.1 & \textbf{87.6} & 75.0 & \textbf{74.2} \\
t-SNE~\cite{tsne} & \textbf{81.1} & \textbf{91.0} & 74.8 & \textbf{44.5} & \textbf{66.0} & \textbf{33.9} & 86.6 & \textbf{78.0} & 73.8 \\
\midrule

 & \multicolumn{3}{c}{\faMicrophone \space UrbanSound} & \multicolumn{3}{c}{\faSatellite \space EuroSAT} & \multicolumn{3}{c}{\faSatellite \space UC Mercered} \\ \cline{2-4}\cline{5-7}\cline{8-10}
\rule{0pt}{2.5ex} & All & Old & New & All & Old & New & All & Old & New \\
 \midrule
 PCA & 32.3 & 36.8 & 44.6 & 37.7 & 46.5 & 56.6 & 37.1 & 55.9 & 40.0 \\
 UMAP & 41.2 & 21.8 & 52.6 & 55.1 & 68.7 & 46.7 & 73.0 & \textbf{76.5} & 74.5 \\
t-SNE & \textbf{45.5} & \textbf{56.7} & \textbf{63.9} & \textbf{68.3} & \textbf{75.5} & \textbf{69.5} & \textbf{75.8} & 67.7 & \textbf{95.7}  \\

\bottomrule
\end{tabular}%
}
\caption{Comparing OmniGCDs performance using the different dimension reduction methods discussed in Section \ref{sec:dim_reduction}. t-SNE achieves the best performance overall. Additional results are available in the supplementary material.}
\label{tab:dim_reduction}
\end{table}


%% file: sec/7_conclusion.tex
\section{Conclusion}
Inspired by human category learning, we have introduced the novel zero-shot GCD problem setting and presented OmniGCD as a modality-agnostic solution. OmniGCD utilizes encoders for each modality to construct a GCD latent space, which represents the GCD task. This space is then transformed into a representation better suited for k-means clustering through the use of our synthetically trained GCDformer. With our evaluation across four modalities and 16 total datasets, we demonstrate how a single OmniGCD model trained only once can perform modality-agnostic GCD without dataset-specific fine-tuning, with improved performance compared to baselines. Taking a modality-agnostic approach illuminates the important role strong encoders play in the GCD process, while inherently decoupling the process of improving GCD task performance from improving the quality of the encoder, allowing improvements in encoder quality to directly translate to improved GCD performance. We encourage future works to investigate methods for improving modality-agnostic performance, as these improvements will propagate to improvements in all modalities.

\section*{Acknowledgements}
This work was supported by the SmartSat CRC (funded by the Australian Government’s CRC Program) and by Shield AI, a global leader in AI pilots for defence and civilian applications.


%% file: supp_sec/supp_intro.tex
\noindent In our supplementary material, we present the following sections in support of the main paper. We first provide qualitative examples showing GCDformer's ability to optimize the GCD latent spaces for k-means clustering in Section~\ref{sec:GCDformer_optim}. We then provide further details on the setup for OmniGCD's zero-shot GCD experiments in Section~\ref{sec:dataset_stats}. Next, in Section~\ref{sec:dimension_ab}, we report ablation results for training GCDformer with higher data dimensionality than the 2D version used for the main results. In Section~\ref{sec:encoder_abs}, we present results using additional encoders for the vision modality, specifically MobileNetV3~\cite{MBv3} and DINOv3~\cite{dinov3}. We also include standard deviation reporting for the results in Table~4 of the main paper in Section~\ref{sec:further_analysis}. Additionally, we provide a more detailed analysis of OmniGCD performance by inspecting the contributions of the dimension reduction method and GCDformer. Section~\ref{sec:vis_enc_ft} contains fine-tuning results for the datasets not presented in Table~5 of the main paper. Lastly, Section~\ref{sec:dim_reduction_method_abs} contains additional results comparing OmniGCD performance across the considered dimension reduction methods.

%% file: supp_sec/GCDFormer_optim.tex
\section{GCDFormer Optimization of GCD Latent Spaces}
\label{sec:GCDformer_optim}

\begin{figure*}
    \centering
    \begin{minipage}{0.49\textwidth}
        \centering
        \begin{subfigure}[b]{\linewidth}
        
            \includegraphics[width=\linewidth]{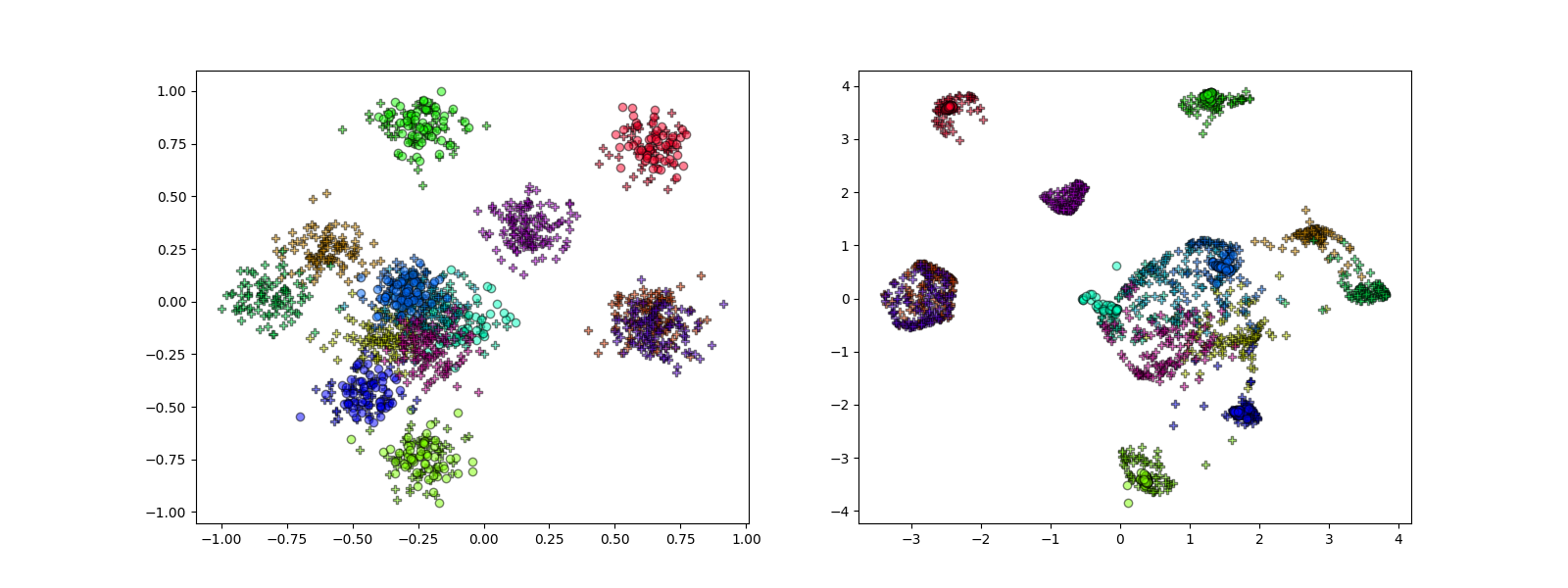}
        \end{subfigure}
        \begin{subfigure}[b]{\linewidth}
            \includegraphics[width=\linewidth]{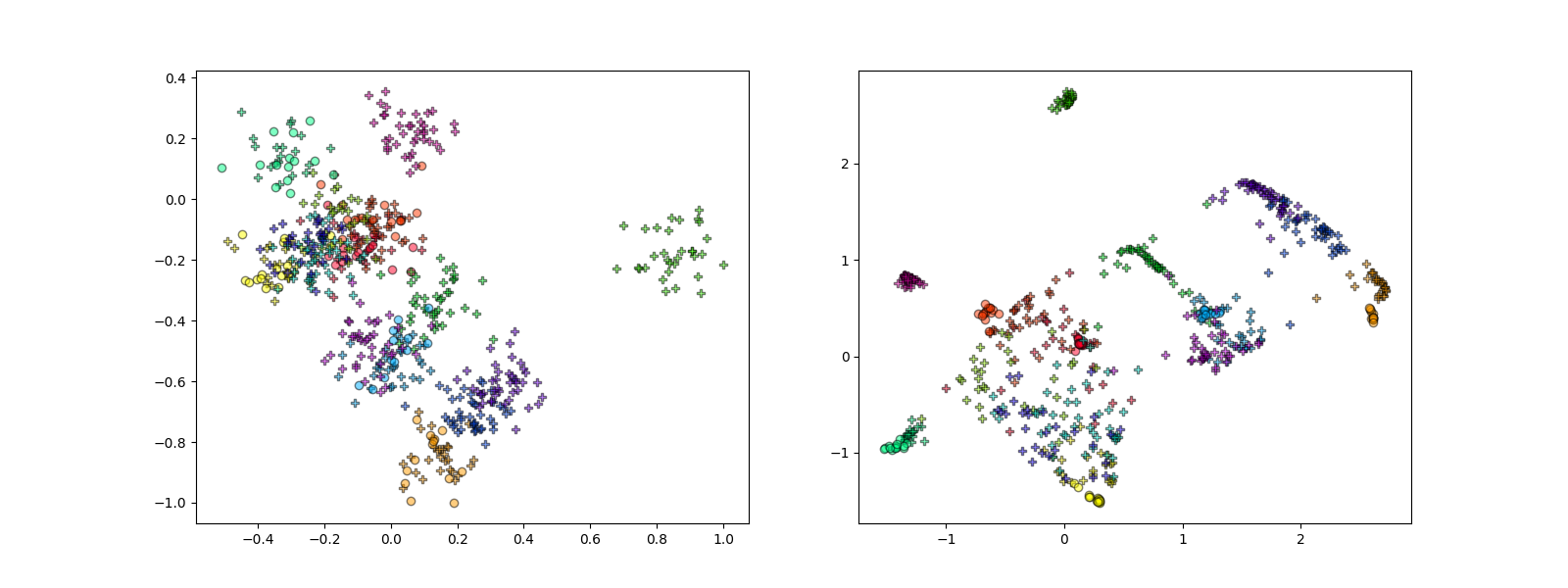}
        \end{subfigure}
        \begin{subfigure}[b]{\linewidth}
        
            \includegraphics[width=\linewidth]{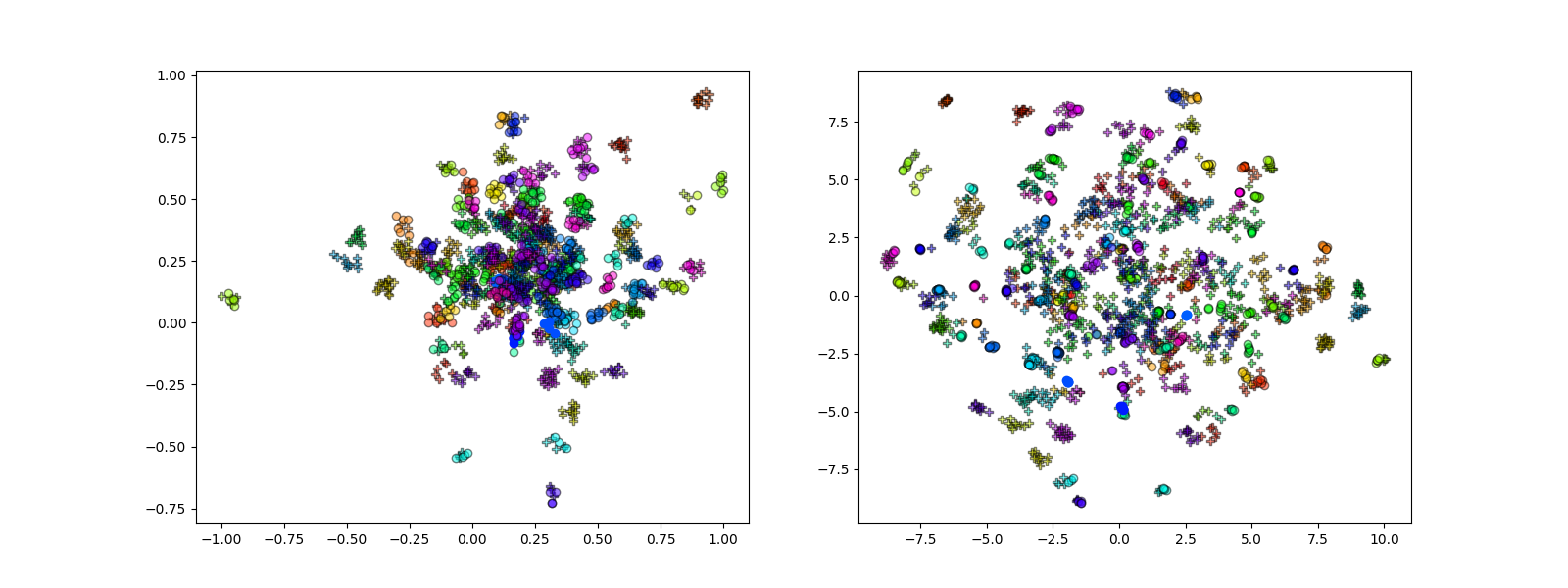}
        \end{subfigure}
        \begin{subfigure}[b]{\linewidth}
            \includegraphics[width=\linewidth]{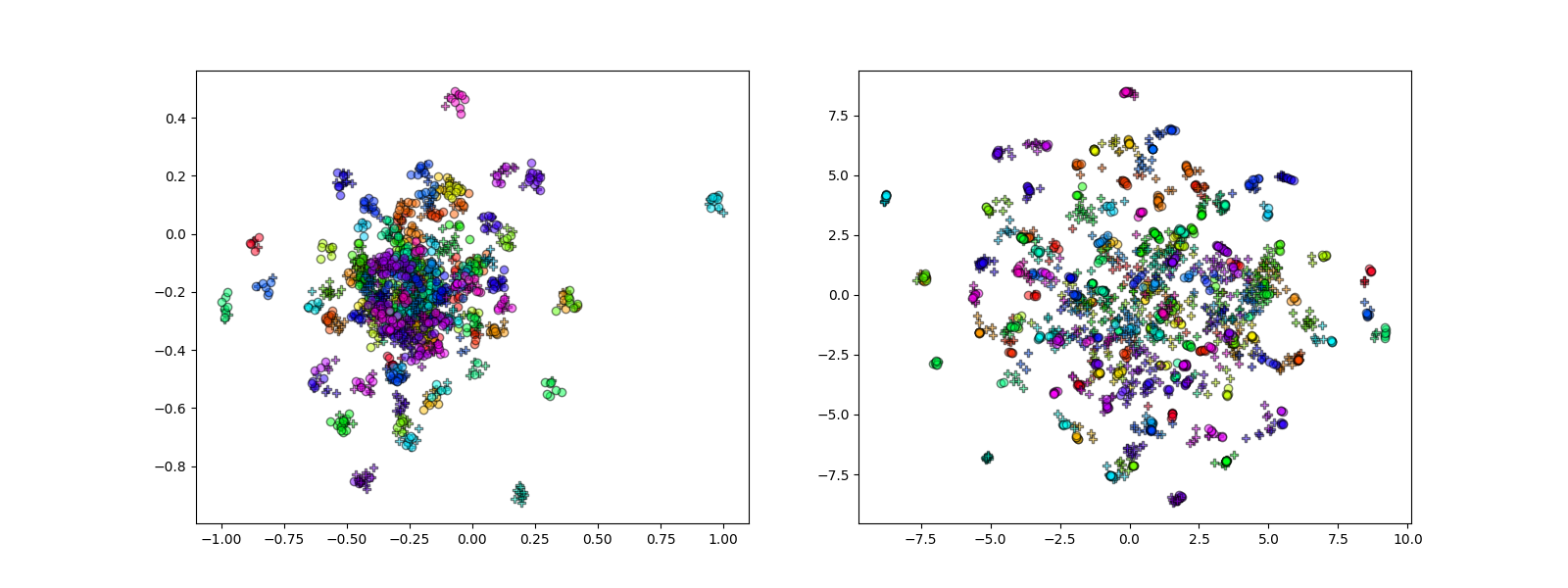}
        \end{subfigure}
        \begin{subfigure}[b]{\linewidth}
            \includegraphics[width=\linewidth]{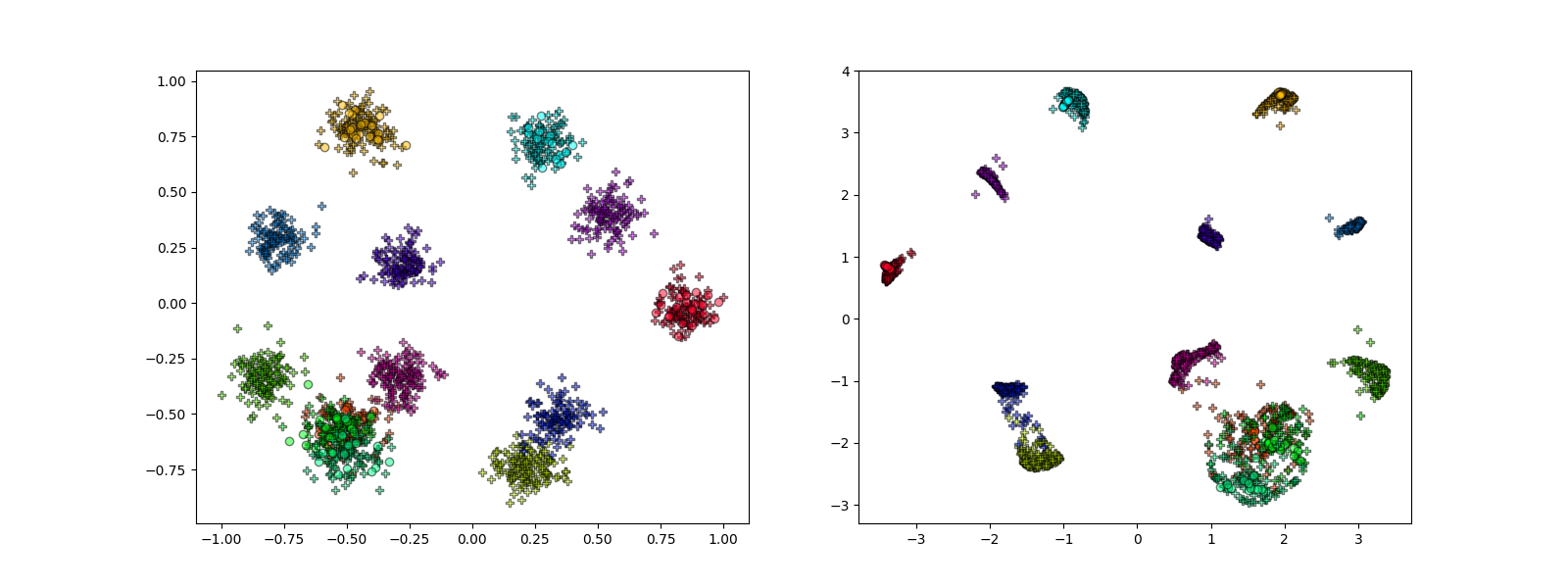}
        \end{subfigure}
        \begin{subfigure}[b]{\linewidth}
        
            \includegraphics[width=\linewidth]{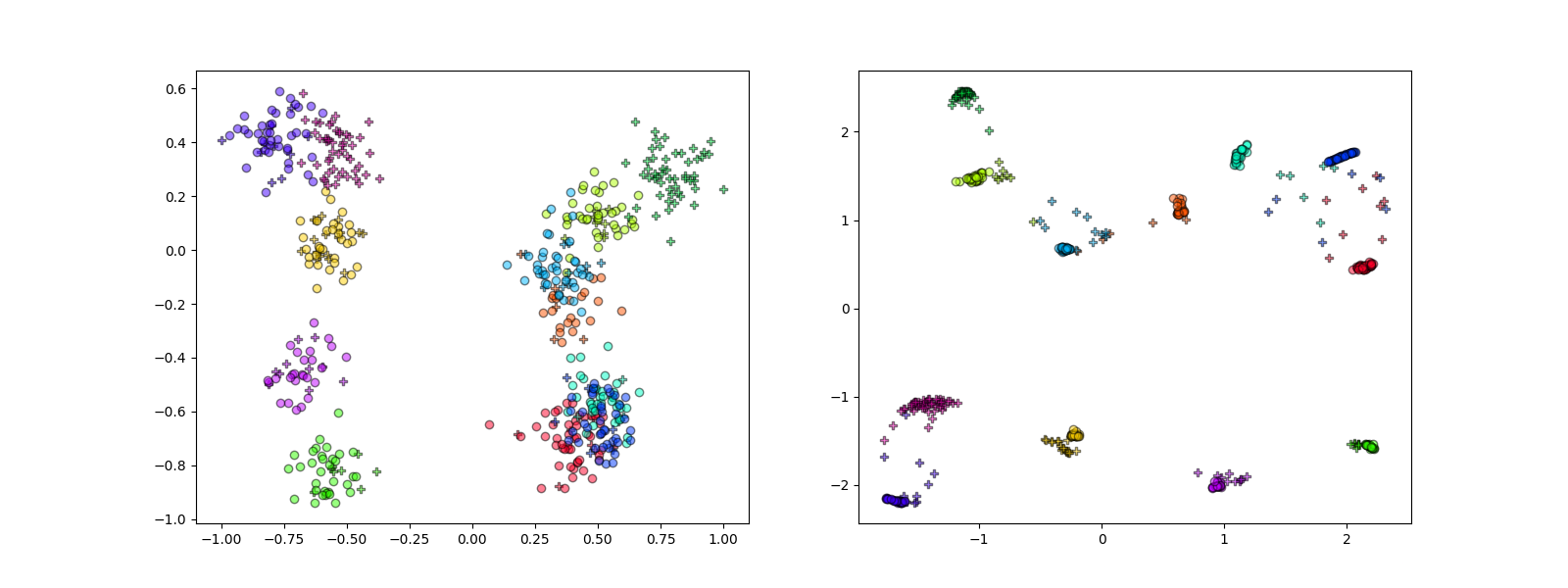}
        \end{subfigure}
    \end{minipage}
    \hfill
    \begin{minipage}{0.49\textwidth}
        \centering
        \begin{subfigure}[b]{\linewidth}
        
            \includegraphics[width=\linewidth]{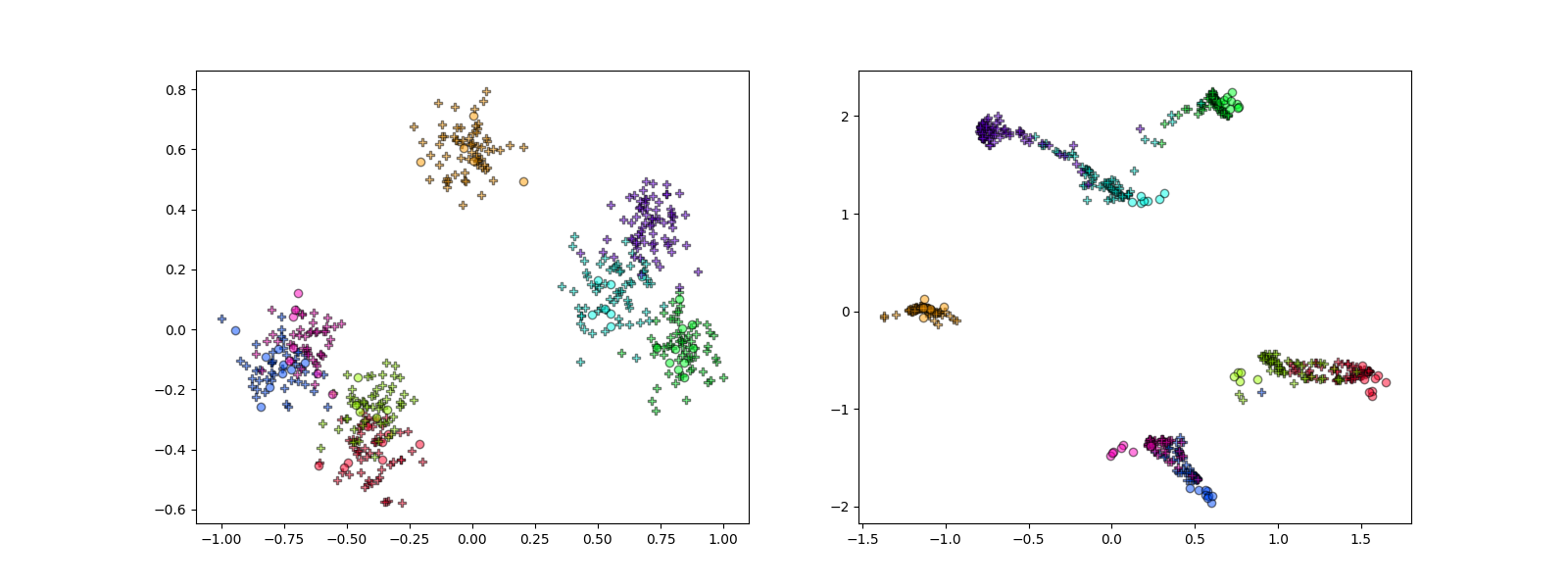}
        \end{subfigure}
        \begin{subfigure}[b]{\linewidth}
        
            \includegraphics[width=\linewidth]{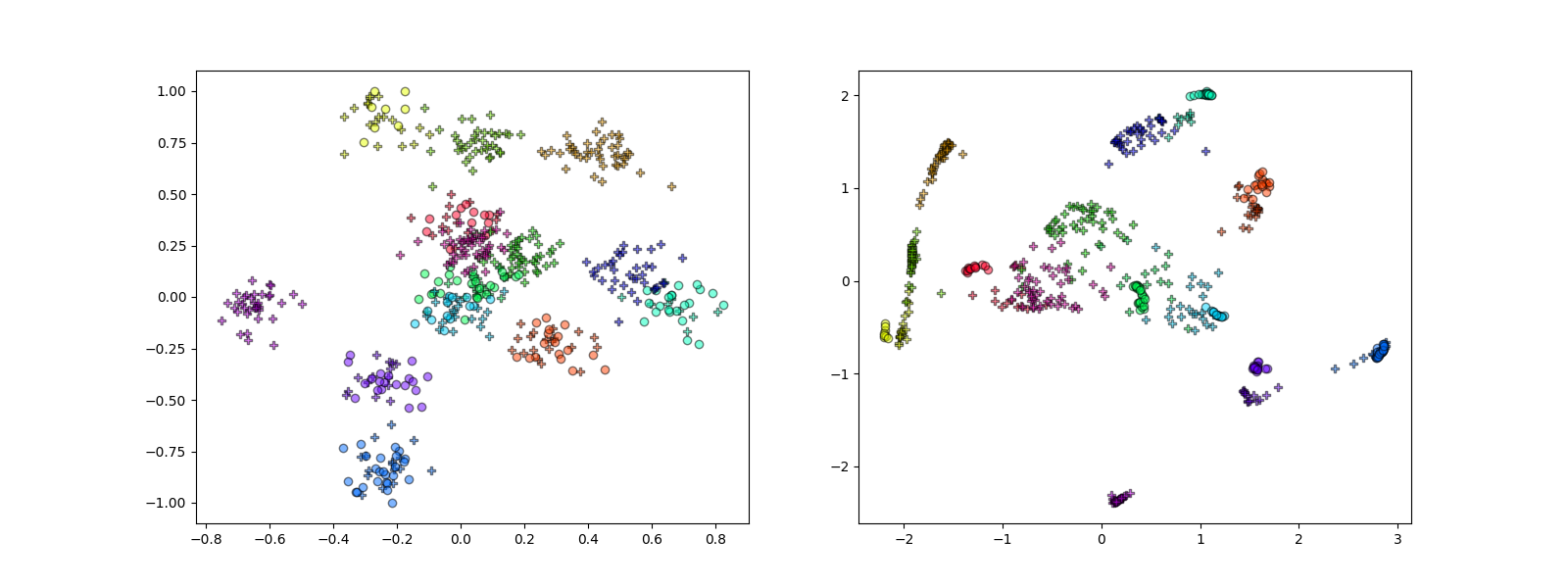}
        \end{subfigure}
        \begin{subfigure}[b]{\linewidth}
        
            \includegraphics[width=\linewidth]{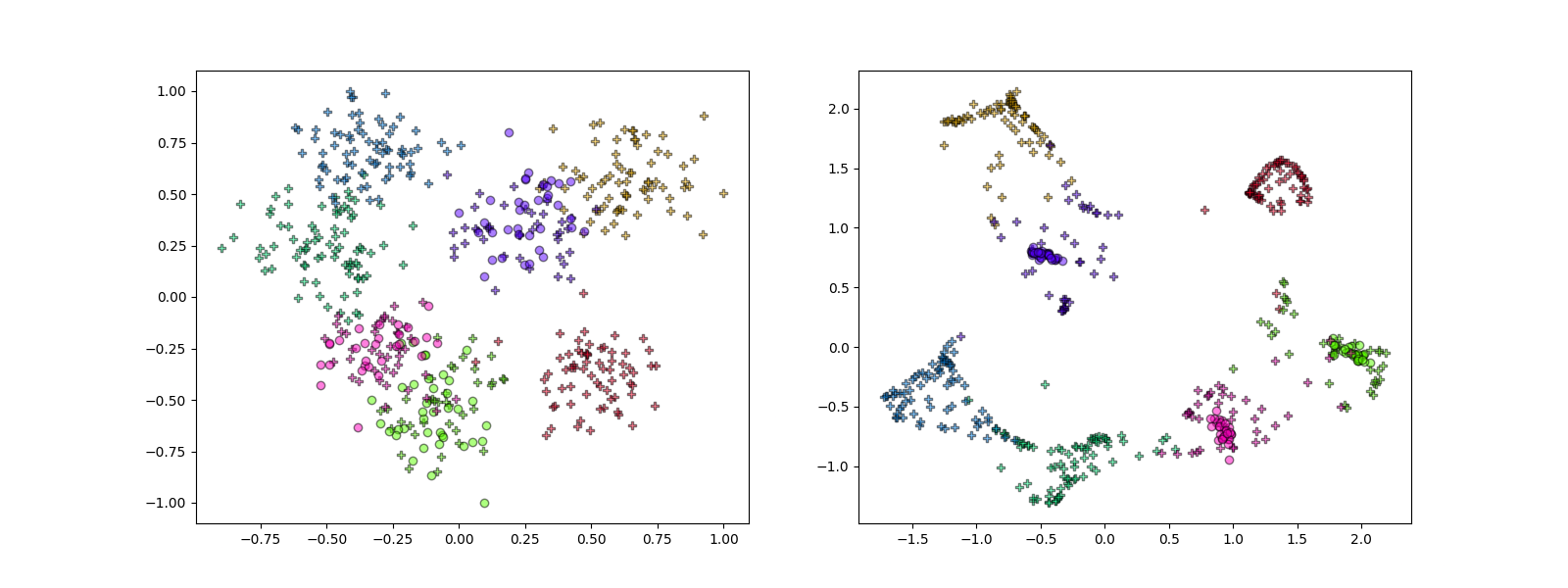}
        \end{subfigure}
        \begin{subfigure}[b]{\linewidth}
            \includegraphics[width=\linewidth]{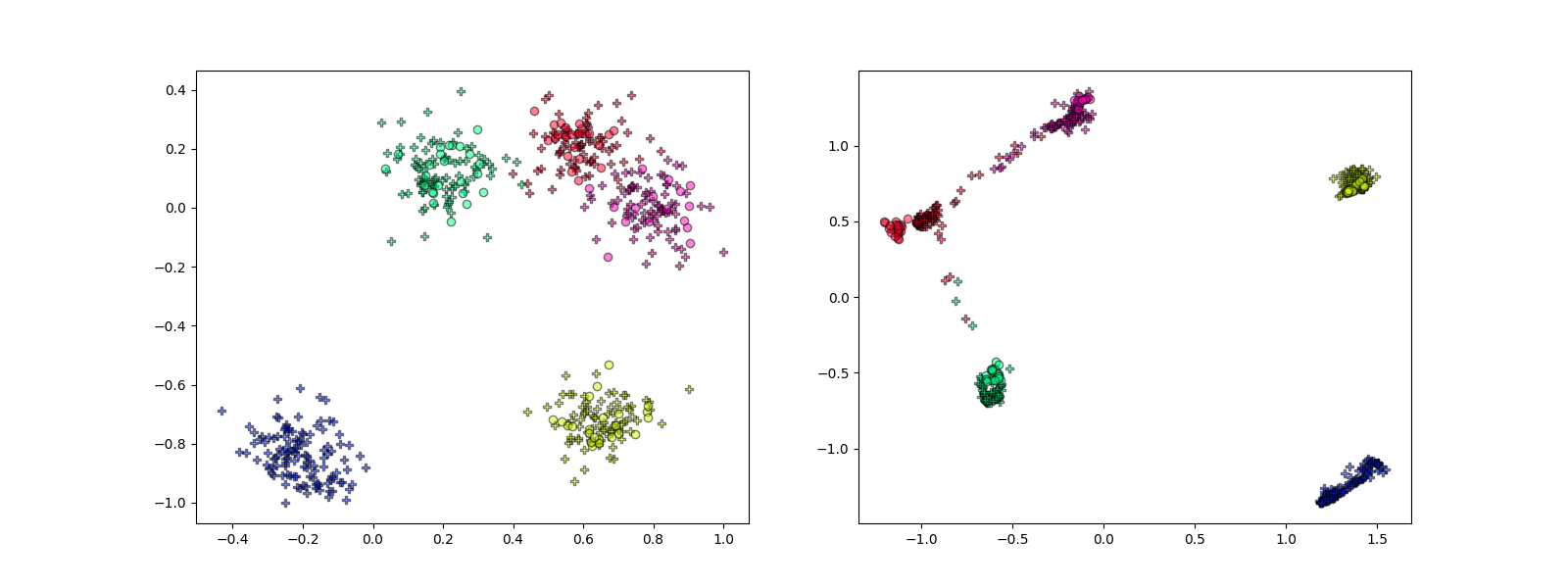}
        \end{subfigure}
        \begin{subfigure}[b]{\linewidth}
        
            \includegraphics[width=\linewidth]{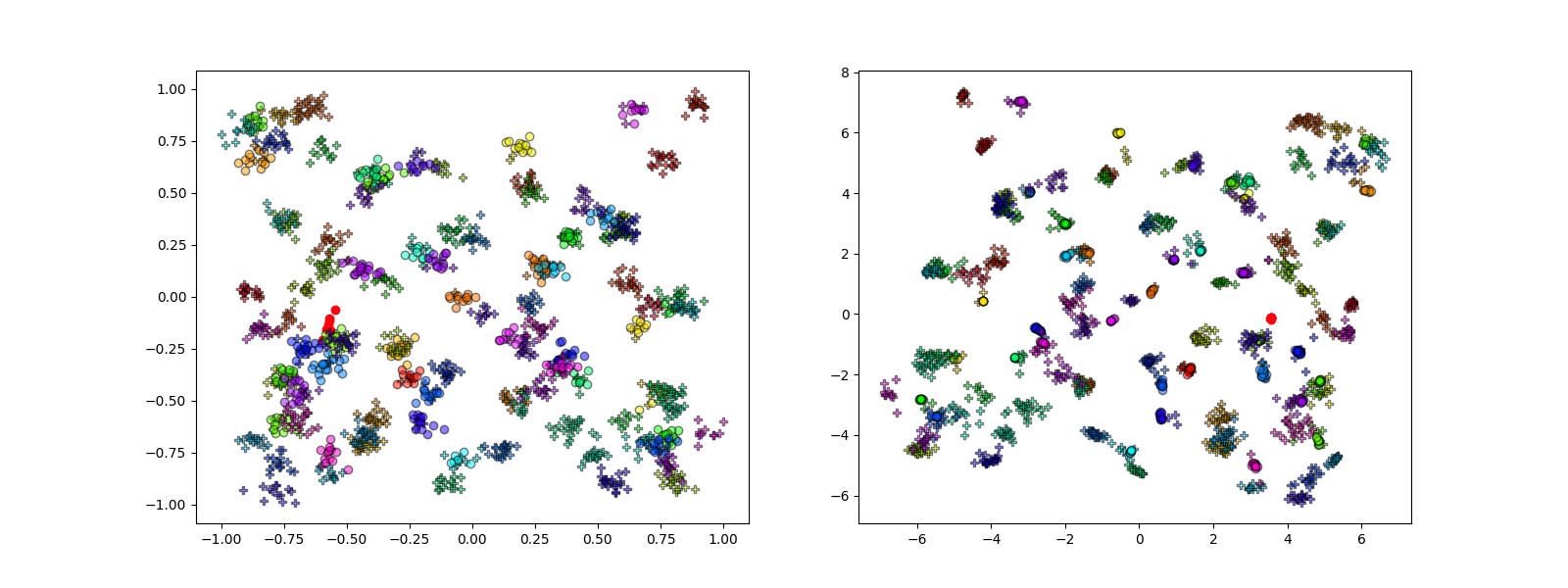}
        \end{subfigure}
        \begin{subfigure}[b]{\linewidth}
        
            \includegraphics[width=\linewidth]{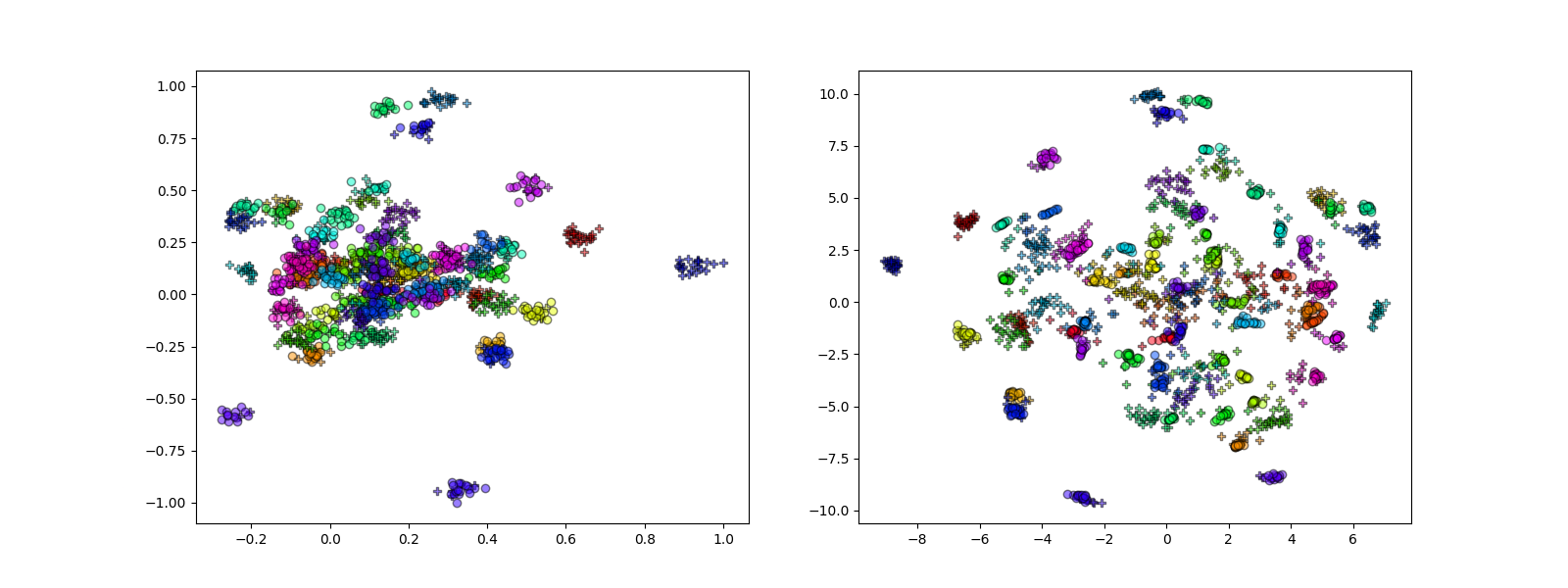}
        \end{subfigure}
    \end{minipage}
    \caption{Qualitative examples of GCDformers ability to optimize the GCD latent spaces for k-means clustering. We show 10 examples of initial (left for each example) and opitmized GCD latent spaces (right for each example). These examples are randomly selected from GCDformers training. In these figures, circular points ($\circ$) represent data belonging to the labeled subset $D_L$, while data belonging to the unknown subnet $D_U$ is represented by a plus sign (\fplus)}
    \label{fig:syn_train_ex}
\end{figure*}

The GCDformer in OmniGCD is trained exclusively on synthetic data to optimize the GCD latent space for clustering. In Figure~\ref{fig:syn_train_ex}, we present qualitative examples illustrating OmniGCD's ability to enhance GCD latent spaces. These examples comprise randomly selected synthetic GCD latent spaces from GCDformer's training. Overall, these visualizations demonstrate GCDformer's ability to optimize the input GCD latent spaces for improved clustering. Moreover, they highlight the adaptability of GCDformer. For instance, in latent spaces with already well-separated clusters, GCDformer tightens the clusters while preserving separation. In contrast, for spaces featuring many overlapping tight clusters, it increases inter-cluster separation. In all cases, GCDformer relies solely on labels from a random subset of points to guide the optimization.

%% file: supp_sec/dataset_stats.tex
\section{Details regarding OmniGCDs Zero-shot setup}
\label{sec:dataset_stats}
In this section, we provide additional details on the sizes of the labeled and unlabeled datasets used by OmniGCD for the main results in Table~4 of the main paper. As noted, at test time, OmniGCD's input comprises both the labeled and unlabeled subsets. In the zero-shot GCD setting, the labeled subset is constructed from training-split samples belonging to the labeled classes, while the unlabeled subset is the test set. During experimentation, we observed that OmniGCD's test-set performance varies slightly with the number of labeled samples per class drawn from the training data. For example, this variation is typically less than one percentage point in \textit{All}, \textit{Old}, and \textit{New} accuracy for $\pm$5 samples per class. Given this variation, we performed hyperparameter sweeps to identify the optimal number of labeled samples per class for each dataset and encoder. These optimal values, used for all reported results, are listed in Table~\ref{tab:support}. We encourage future work to further examine the impact of the number of labeled samples per class, as a key desirable trait of a modality-agnostic GCD method is improved performance with additional samples per class.

\begin{table}[t]
\resizebox{\linewidth}{!}{%
\begin{tabular}{llccc}\toprule
Dataset & Encoder & Labeled Data per Class & Total Labeled Data & Total Unlabeled Data  \\
\midrule
\multirow{2}{*}{\faEye[regular] CIFAR10} & DINOv1 & 90 & 450 & 10,000  \\
 & DINOv2 & 90 & 450 & 10,000  \\
\multirow{2}{*}{\faEye[regular] CIFAR100} & DINOv1 & 150 & 12,000 & 10,000  \\
 & DINOv2 & 30 & 2,400 & 10,000  \\
\multirow{2}{*}{\faEye[regular] IN100} & DINOv1 & 20 & 1,000 & 5,000  \\
 & DINOv2 & 10 & 500 & 5,000 \\
 \multirow{2}{*}{\faEye[regular] CUB-200} & DINOv1 & 25 & 2,500 & 5,794 \\
 & DINOv2 & 25 & 2,500 & 5,794\\
 \multirow{2}{*}{\faEye[regular] SCars} & DINOv1 & 45 & 3,988 & 8,041 \\
 & DINOv2 & 35 & 3,390 & 8,041\\
 \multirow{2}{*}{\faEye[regular] Aircraft} & DINOv1 & 20 & 1,000 & 3,333 \\
 & DINOv2 & 15 & 750 & 3,333 \\
 \multirow{2}{*}{\faEye[regular] Herb-19} & DINOv1 & 5 & 1,705 & 2,697 \\
 & DINOv2 & 20 & 6,506 & 2,697 \\
 \midrule
 \faFile*[regular] BANKING & e5-Large & 80 & 2,967 & 3,080 \\
 \faFile*[regular] StackOverflow & e5-Large & 40 & 400 & 1,000 \\
 \faFile*[regular] CLINIC & e5-Large & 10 & 750 & 3,000\\
 \midrule
 \faMicrophone VocalSet & MERT-95M & 60 & 478 & 2,285\\
 \faMicrophone UrbanSound & MERT-95M & 60 & 300 & 4,366 \\
 \midrule
 \faSatellite EuroSAT & DOFA-B & 50 & 250 & 5,400 \\
 \faSatellite So2SAT & DOFA-B & 20 & 140 & 4,838\\
 \faSatellite RESIC45 & DOFA-B & 80 & 1,200 & 6,300\\
 \faSatellite UC Merced & DOFA-B & 50 & 446 & 420 \\
 
 \bottomrule
\end{tabular}%
}
\caption{Number of labeled data per class, total labeled data and total unlabeled data used for optimal results of our GCD Transformer. These ratios were used for results presented in Table 4 of the main paper.}
\label{tab:support}
\end{table}

%% file: supp_sec/OmniGCD_dim_ablation.tex
\section{OmniGCD Dimension Ablations}
\label{sec:dimension_ab}
In the main paper, we chose 2D data for GCDformer training, as low dimensionality ensures tractability in generating training data. In this section, we present ablations on training with higher-dimensional data (specifically, 32, 64, and 128 dimensions). Due to the $O(N^2 k)$ time complexity of t-SNE~\cite{jung2024review}---where $N$ is the number of data points and $k$ the number of dimensions---we conduct only limited ablations on the vision modality using DINOv1~\cite{dinov1}. The results in Table~\ref{tab:OmniGCD_dims} show that OmniGCD's overall performance degrades as the input dimensionality increases. This validates our design choice of 2D data for GCDformer.

\begin{table*}[]
\resizebox{\linewidth}{!}{%
\begin{tabular}{@{\extracolsep{4pt}}cccccccccccccccccccccc}
\toprule
   & \multicolumn{3}{c}{\faEye[regular] \space CIFAR-10} & \multicolumn{3}{c}{\faEye[regular] \space CIFAR-100} & \multicolumn{3}{c}{\faEye[regular] \space ImageNet-100} & \multicolumn{3}{c}{\faEye[regular] \space CUB-200} & \multicolumn{3}{c}{\faEye[regular] \space Stanford Cars} & \multicolumn{3}{c}{\faEye[regular] \space Aircraft} & \multicolumn{3}{c}{\faEye[regular] \space Herbarium-19} \\
 \cline{2-4} \cline{5-7} \cline{8-10} \cline{11-13} \cline{14-16} \cline{17-19} \cline{20-22}
Num Dimensions & All & Old & New & All & Old & New & All & Old & New & All & Old & New & All & Old & New & All & Old & New & All & Old & New \\
 \midrule
 2 & \textbf{90.7} & \textbf{97.0} & 91.6 & \textbf{60.0} & \textbf{89.9} & \textbf{74.6} & \textbf{81.1} & \textbf{91.0} & \textbf{74.8} & \textbf{44.5} & \textbf{66.0} & 33.9 & \textbf{12.6} & \textbf{19.6} & \textbf{16.9} & \textbf{18.9} & \textbf{14.3} & \textbf{13.9} & \textbf{30.8} & \textbf{48.6} & \textbf{50.0}\\
32 & 87.2 & 94.6 & 91.0 & 58.5 & 70.8 & 73.8 & 78.5 & 90.0 & 74.2 & 35.9 & 58.0 & \textbf{45.4} & \textbf{12.6} & 13.7 & 16.3 & 13.2 & 9.0 & 10.3 & 26.8 & 37.1 & 45.0 \\
64 & 51.7 & 67.6 & 27.9 & 56.9 & 71.5 & 69.4 & 61.8 & 78.0 & 46.4 & 28.9 & 43.3 & 26.8 & 12.3 & 12.8 & 14.9 & 12.9 & 13.1 & 7.9 & 27.2 & 42.9 & \textbf{50.0} \\
128 & 74.7 & 95.8 & \textbf{91.7} & 39.7 & 82.1 & 57.1 & 65.1 & 86.0 & 43.8 & 23.9 & 54.0 & 45.1 & 8.1 & 4.8 & 6.1 & 12.3 & 11.3 & 9.1 & 22.8 & 42.9 & \textbf{50.0}
   \\
 \bottomrule
\end{tabular}%
}
\caption{Ablation of zero-shot GCD performance for OmniGCD using the DINOv1 (ViT-B/16)~\cite{dinov1} vision encoder. Features from DINOv1 are projected to varying t-SNE embedding dimensions (2, 32, 64, 128) across the vision modality datasets. We \textbf{bold} the best results for each comparison.}
\label{tab:OmniGCD_dims}
\end{table*}

%% file: supp_sec/additional_encoders.tex
\section{Additional Vision Encoders Ablations}
\label{sec:encoder_abs}
Here, we present results using two additional pretrained vision encoders: MobileNetV3~\cite{MBv3} and DINOv3~\cite{dinov3}. These results are not merely to demonstrate that OmniGCD works with additional encoders, as the main results with multimodal encoders already establish this, but serve two specific purposes: (1) MobileNetV3 is a lightweight encoder suitable for edge devices, and we aim to demonstrate OmniGCD's performance with such an encoder; (2) DINOv3 is the most recent and highest-performing encoder in the DINO family, and we provide these results for future GCD studies that will likely adopt this encoder. As shown in Table~\ref{tab:additional_encoders}, with MobileNetV3, OmniGCD achieves the best \textit{All} accuracy on 5 of the 7 datasets, the best \textit{Old} accuracy on 4 of the 7 datasets, and the best \textit{New} accuracy on 3 of the 7 datasets. With DINOv3, it achieves the best \textit{All} accuracy on all 7 datasets, the best \textit{Old} accuracy on 3 of the 7 datasets, and the best \textit{New} accuracy on 3 of the 7 datasets.

\begin{table*}[]
\resizebox{\linewidth}{!}{%
\begin{tabular}{@{\extracolsep{4pt}}llccccccccccccccccccccc}
\toprule
 &  & \multicolumn{3}{c}{\faEye[regular] \space CIFAR-10} & \multicolumn{3}{c}{\faEye[regular] \space CIFAR-100} & \multicolumn{3}{c}{\faEye[regular] \space ImageNet-100} & \multicolumn{3}{c}{\faEye[regular] \space CUB-200} & \multicolumn{3}{c}{\faEye[regular] \space Stanford Cars} & \multicolumn{3}{c}{\faEye[regular] \space Aircraft} & \multicolumn{3}{c}{\faEye[regular] \space Herbarium-19} \\
 \cline{3-5} \cline{6-8} \cline{9-11} \cline{12-14} \cline{15-17} \cline{18-20} \cline{21-23}
 &  & All & Old & New & All & Old & New & All & Old & New & All & Old & New & All & Old & New & All & Old & New & All & Old & New \\
 \midrule
\multirow{3}{*}{MobileNetV3} & K-means & 66.4 & {  74.7} & {  71.0} & 35.6 & 16.5 & {  41.2} & {  69.5} & {  59.9} & 64.7 & {  45.5} & 34.5 & {  45.2} & 14.2 & {  34.9} & {  8.8} & 15.3 & {  14.8} & 16.2 & 25.8 & 12.9 & \textbf{18.3} \\
 & GCD (w/o FT) & {  71.2} & \textbf{89.0} & \textbf{82.0} & \textbf{48.4} & \textbf{48.5} & 26.5 & 56.6 & 46.0 & {  76.0} & 44.4 & {  50.0} & 38.3 & \textbf{17.8} & \textbf{64.4} & \textbf{11.9} & {  16.9} & 6.0 & \textbf{19.7} & {  26.4} & \textbf{14.3} & 11.1 \\
 & OmniGCD & \textbf{77.5} & 72.5 & 70.6 & {  40.2} & {  23.5} & \textbf{44.0} & \textbf{77.1} & \textbf{61.0} & \textbf{84.0} & \textbf{54.2} & \textbf{53.0} & \textbf{53.7} & {  16.4} & 23.0 & 7.4 & \textbf{17.2} & \textbf{15.8} & {  18.8} & \textbf{28.3} & \textbf{14.3} & {  16.7} \\
 \midrule
DINOv3 & K-means & {  86.1} & 67.5 & {  95.6} & 66.6 & {  49.7} & {  96.3} & {  78.2} & 27.8 & 83.8 & {  71.3} & 76.7 & {  58.8} & {  58.0} & {  60.6} & {  72.6} & {  39.0} & {  38.8} & 59.0 & 30.2 & \textbf{34.9} & {  32.3} \\
 & GCD (w/o FT) & 80.5 & {  76.6} & 95.5 & {  70.6} & \textbf{72.5} & \textbf{97.0} & 67.0 & {  36.0} & \textbf{88.0} & 63.6 & \textbf{95.0} & 29.9 & 56.4 & 45.0 & 46.7 & 36.5 & \textbf{41.8} & \textbf{62.7} & {  33.5} & {  33.3} & \textbf{50.0} \\
 & OmniGCD & \textbf{97.2} & \textbf{95.1} & \textbf{98.8} & \textbf{74.7} & 40.1 & 93.4 & \textbf{88.2} & \textbf{38.0} & {  85.0} & \textbf{79.3} & {  84.7} & \textbf{86.7} & \textbf{68.1} & \textbf{83.0} & \textbf{89.1} & \textbf{44.0} & 36.1 & {  59.7} & \textbf{36.0} & {  33.3} & 25.0 \\
 \bottomrule
\end{tabular}%
}
\caption{Additional Zero-shot GCD results for the vision modality using the MobileNetV3 (MBV3-Large)~\cite{MBv3} and DINOv3 (ViT-B/16)~\cite{dinov3} encoders. We \textbf{bold} the best results for each comparison.}
\label{tab:additional_encoders}
\end{table*}

\begin{table*}[]
\resizebox{\linewidth}{!}{%
\begin{tabular}{@{\extracolsep{4pt}}lccccccccccccccccccccc}
\toprule
  & \multicolumn{3}{c}{\faEye[regular] \space CIFAR-10} & \multicolumn{3}{c}{\faEye[regular] \space CIFAR-100} & \multicolumn{3}{c}{\faEye[regular] \space ImageNet-100} & \multicolumn{3}{c}{\faEye[regular] \space CUB-200} & \multicolumn{3}{c}{\faEye[regular] \space Stanford Cars} & \multicolumn{3}{c}{\faEye[regular] \space Aircraft} & \multicolumn{3}{c}{\faEye[regular] \space Herbarium-19} \\
 \cline{2-4} \cline{5-7} \cline{8-10} \cline{11-13} \cline{14-16} \cline{17-19} \cline{20-22}
  & All & Old & New & All & Old & New & All & Old & New & All & Old & New & All & Old & New & All & Old & New & All & Old & New \\
 \midrule
 GCD & 81.8 & 86.2 & 76.9 & 69.0 & 77.4 & 62.0 & 73.5 & \textbf{92.6} & 63.9 & 51.3 & 56.6 & 48.7 & 39.0 & 57.6 & 29.9 & 45.0 & 41.1 & 46.9 & \textbf{35.4} & 51.0 & 27.0 \\
 OmniGCD (w/o FT) & 90.7 & \textbf{97.0} & 91.6 & 60.0 & \textbf{89.9} & \textbf{74.6} & \textbf{81.1} & 91.0 & 74.8 & 44.5 & 66.0 & 33.9 & 12.6 & 19.6 & 16.9 & 18.9 & 14.3 & 13.9 & 30.8 & 48.6 & \textbf{50.0} \\
 \midrule
 OmniGCD (w FT) & \textbf{96.1} & 96.2 & \textbf{96.1} & \textbf{71.7} & 76.8 & 66.0 & 76.2 & 91.6 & \textbf{94.8} & \textbf{59.8} & \textbf{71.0} & \textbf{90.6} & \textbf{49.1} & \textbf{63.7} & \textbf{48.1} & \textbf{47.5} & \textbf{71.21} & \textbf{59.4} & 34.4 & \textbf{79.2} & 39.6 \\
 \bottomrule
\end{tabular}%
}
\caption{Results on the standard GCD setting compared to the original GCD method which fine-tunes its encoder~\cite{GCD2022}. We fine-tune the vision encoder using the same way as the original GCD method~\cite{GCD2022}. We \textbf{bold} the best results for each comparison.}
\label{tab:additional_FT}
\end{table*}

%% file: supp_sec/tsne_vs_gcdformer.tex
\section{Further analysis of OmniGCD Zero-shot GCD performance}
\label{sec:further_analysis}
In this section, we provide a more granular analysis of OmniGCD's performance, along with standard deviations for the main results in Table~4 of the main paper. These results are presented in Table~\ref{tab:further_analysis}. For the granular analysis, we compare the individual contributions of t-SNE dimensionality reduction and GCDformer. Using neither component yields k-means clustering directly on the feature vectors, while using both yields the full OmniGCD method. We omit the variant using GCDformer without t-SNE, as it would require training separate GCDformer models per modality. Overall, combining t-SNE and GCDformer improves clustering performance over t-SNE alone in 80\% of measured metrics across all datasets. The sole exception is \textit{New} accuracy on the Aircraft~\cite{aircraft} dataset with DINOv2~\cite{dinov2}, where the combination worsens performance. In 5 instances, GCDformer fails to further improve clustering over t-SNE alone, and in 6 instances, it degrades performance. Notably, we observe the counterintuitive yet consistent improvements in GCD performance from t-SNE alone, despite the information loss from reducing feature vectors to 2D. We would expect such dimensionality reduction to impair clustering, but it does not. This suggests that dimensionality reduction is a valuable tool for GCD methods. It represents an intriguing discovery, hinting at an unintentional alignment between dimensionality reduction and GCD tasks. We encourage future work to explore this further, as exploiting such alignment could enhance GCD performance in subsequent methods.

\begin{landscape}
\begin{table}[t]
\resizebox{\linewidth}{!}{%
\begin{tabular}{@{\extracolsep{4pt}}lllccccccccccccccccccccc}
\toprule
&  & \multirow{3}{*}{\rotatebox{90}{GCDformer}}
\\  \\ & \multirow{2}{*}{\rotatebox{90}{t-SNE}}  

 &  &  \multicolumn{3}{c}{\faEye[regular] \space CIFAR-10} & \multicolumn{3}{c}{\faEye[regular] \space CIFAR-100} & \multicolumn{3}{c}{\faEye[regular] \space ImageNet-100} & \multicolumn{3}{c}{\faEye[regular] \space CUB-200} & \multicolumn{3}{c}{\faEye[regular] \space Stanford Cars} & \multicolumn{3}{c}{\faEye[regular] \space Aircraft} & \multicolumn{3}{c}{\faEye[regular] \space Herbarium-19} \\ 
 
 \cline{4-6}\cline{7-9}\cline{10-12}\cline{13-15}\cline{16-18}\cline{19-21}\cline{22-24}
 
\rule{0pt}{11pt} &  &  & All & Old & New & All & Old & New & All & Old & New & All & Old & New & All & Old & New & All & Old & New & All & Old & New \\ \midrule

\multirow{2}{*}{\rule{0pt}{2ex} \rotatebox{90}{\scriptsize Dv1~\cite{dinov1}}} & \faTimes & \faTimes & 77.8 ± 6.6 & 88.1 ± 2.2 & 60.3 ± 38.6 & 52.5 ± 0.6 & 68.9 ± 12.1 & 66.3 ± 5.8 & 73.5 ± 1.3 & 87.6 ± 2.5 & 69.9 ± 17.4 & 35.8 ± 1.0 & 54.7 ± 9.3 & 39.8 ± 9.4 & 10.7 ± 0.2 & 14.5 ± 2.5 & 15.9 ± 1.7 & 14.8 ± 0.7 & 10.9 ± 4.6 & 10.8 ± 5.2 & 24.9 ± 0.4 & 42.9 ± 10.7 & 22.5 ± 11.5\\

& \faCheck & \faTimes & 83.6 ± 6.3 & 91.5 ± 10.1 & 85.6 ± 13.1 & 57.7 ± 0.8 & 79.1 ± 8.4 & 74.5 ± 2.9 & 78.4 ± 1.2 & 89.8 ± 2.4 & \textbf{74.8} ± 0.4 & 44.2 ± 0.5 & 64.0 ± 3.3 & \textbf{38.6} ± 5.6 & \textbf{12.6} ± 0.1 & 10.8 ± 2.8 & 13.0 ± 0.5 & 18.2 ± 0.2 & 11.3 ± 1.5 & \textbf{17.6} ± 7.2 & 30.2 ± 0.2 & 42.9 ± 0.0 & 40.0 ± 12.2 \\

 & \faCheck & \faCheck & \textbf{90.7} ± 3.9 & \textbf{97.0} ± 3.1  & \textbf{91.6} ± 0.4  & \textbf{60.0} ± 0.8 & \textbf{89.9} ± 7.4 & \textbf{74.6} ± 16.9 & \textbf{81.1} ± 1.5 & \textbf{91.0} ± 0.0 & \textbf{74.8} ± 0.5 & \textbf{44.5} ± 1.5 & \textbf{66.0} ± 2.7 & 33.9 ± 6.6 & \textbf{12.6} ± 0.2 & \textbf{19.6} ± 13.6 & \textbf{16.9} ± 3.6 & \textbf{18.9} ± 0.8 & \textbf{14.3} ± 3.3  & 13.9 ± 6.4 & \textbf{30.8} ± 0.2 & \textbf{48.6} ± 5.7 & \textbf{50.0} ± 0.0 \\
 \midrule
 
\multirow{2}{*}{\rule{0pt}{2ex} \rotatebox{90}{\scriptsize Dv2~\cite{dinov2}}} & \faTimes & \faTimes & 82.1 ± 8.5 & 62.6 ± 25.4 & 95.0 ± 4.3 & 69.1 ± 1.8  & 44.2 ± 20.4 & 51.3 ± 6.4 & 79.9 ± 1.3 & 87.7 ± 1.7 & 78.4 ± 0.7 & 70.3 ± 0.9 & 95.8 ± 4.9 & 75.4 ± & 27.3 ± 0.6 & 16.8 ± 6.7 & 28.7 ± 3.8 & 19.6 ± 0.3 & 12.7 ± 4.9 & \textbf{22.5} ± 3.9 & 29.0 ± 0.5 & 40.8 ± 5.1 & 23.2 ± 3.3 \\

& \faCheck & \faTimes & 89.9 ± 4.6 & 93.3 ± 8.6 & 91.0 ± 7.8 & 75.2 ± 0.7 & \textbf{48.5} ± 8.6 & 47.0 ± 8.0 & 86.8 ± 1.3 & 84.4 ± 18.2 & \textbf{90.0} ± 0.0 & 77.5 ± 0.3 & 99.3 ± 1.3 & 79.29 ± 21.0 & 33.3 ± 0.4 & 23.2 ± 3.4 & 38.4 ± 6.7 & \textbf{21.2} ± 0.2 & 15.8 ± 1.2 & 18.2 ± 6.9 & 33.5 ± 0.2 & \textbf{60.0} ± 3.1 & \textbf{35.5} ± 4.1  \\

 & \faCheck & \faCheck & \textbf{96.9} ± 3.5 & \textbf{96.9} ± 0.7 & \textbf{95.6} ± 0.4 & \textbf{78.1} ± 1.3 & 47.9 ± 4.5 & \textbf{56.8} ± 9.6 & \textbf{88.7} ± 0.4 & \textbf{94.0} ± 0.0 & \textbf{90.0} ± 0.0 & \textbf{79.8} ± 0.9 & \textbf{100.0} ± 0.0 & \textbf{96.4} ± 0.0 & \textbf{33.4} ± 0.6 & \textbf{24.2} ± 2.8 & \textbf{43.1} ± 1.1 & \textbf{21.2} ± 0.1 & \textbf{17.0} ± 2.2 & 11.6 ± 7.4 & \textbf{34.8} ± 0.5 & \textbf{60.0} ± 1.5 & 25.8 ± 9.7 \\
 \midrule
 &  & & \multicolumn{3}{c}{\faFile*[regular] BANKING} & \multicolumn{3}{c}{\faFile*[regular] StackOverflow} & \multicolumn{3}{c}{\faFile*[regular] CLINIC} & \multicolumn{3}{c}{\faMicrophone \space VocalSet} & \multicolumn{3}{c}{\faMicrophone \space UrbanSound} &  \multicolumn{3}{c}{\faSatellite \space EuroSAT} & \multicolumn{3}{c}{\faSatellite \space So2SAT} \\ \cline{4-6}\cline{7-9}\cline{10-12}\cline{13-15}\cline{16-18}\cline{19-21}\cline{22-24}
\rule{0pt}{11pt} & & & All & Old & New & All & Old & New & All & Old & New & All & Old & New & All & Old & New & All & Old & New & All & Old & New \\ \midrule

 & \faTimes & \faTimes & 57.7 ± 1.5 & 64.4 ± 14.3 & 43.4 ± 8.2 & 72.9 ± 3.7 & 68.3 ± 11.8 & 45.2 ± 9.3 & 69.5 ± 1.6 & 85.0 ± 13.4 & 81.7 ± 16.0 & 22.1 ± 1.5 & 15.0 ± 2.7 & 25.0 ± 6.0 & 39.3 ± 2.0 & 37.6 ± 10.6 & 37.7 ± 11.7 & 53.2 ± 2.4 & 38.4 ± 6.2 & 62.4 ± 7.1 & 27.3 ± 1.1 & 32.5 ± 2.3 & 65.1 ± 1.1 \\
 
 & \faCheck & \faTimes & 65.2 ± 1.2 & 41.3 ± 6.0 & \textbf{51.1} ± 10.0 & 84.8 ± 3.1 & \textbf{90.2} ± 6.6 & \textbf{85.4} ± 5.4 & 81.4 ± 0.8 & 85.0 ± 6.4 & 72.7 ± 11.3 & 22.1 ± 1.3 & 14.7 ± 1.3 & 32.7 ± 7.0 & 44.8 ± 2.4 & 53.8 ± 5.8 & 52.5 ± 10.3 & 66.0 ± 2.5 & 71.1 ± 2.3 & 61.0 ± 14.5 & 31.1 ± 1.8 & \textbf{39.5} ± 3.6 & 48.2 ± 5.9 \\

 & \faCheck & \faCheck & \textbf{66.4} ± 2.2 & \textbf{77.1} ± 34.8 & 40.8 ± 17.8 & \textbf{86.8} ± 2.0 & 78.0 ± 15.2 & 73.8 ± 12.3 & \textbf{82.1} ± 1.6 & \textbf{96.0} ± 5.6 & \textbf{90.0} ± 12.1 & \textbf{25.6} ± 2.3 & \textbf{17.5} ± 4.7 & \textbf{49.2} ± 5.8 & \textbf{45.5} ± 1.0 & \textbf{56.7} ± 9.7 & \textbf{63.9} ± 5.3 & \textbf{68.3} ± 6.3 & \textbf{75.5} ± 0.8 & \textbf{69.5} ± 19.0 & \textbf{31.7} ± 0.5 & 38.4 ± 2.8 & \textbf{58.4} ± 2.9  \\
 
 \midrule
 &  & & \multicolumn{3}{c}{\faSatellite \space RESIC45} & \multicolumn{3}{c}{\faSatellite \space UC Merced} \\\cline{4-6}\cline{7-9}
\rule{0pt}{11pt}  & & & All & Old & New & All & Old & New  \\ \cline{4-9}
\rule{0pt}{11pt} & \faTimes & \faTimes & 42.8 ± 2.2 & 37.1 ± 7.7 & 60.8 ± 8.6 & 63.6 ± 3.0 & 50.6 ± 10.4 & 70.8 ± 12.9 \\

& \faCheck & \faTimes & 55.6 ± 2.5 & 60.4 ± 19.4 & 71.3 ± 7.7 & 64.1 ± 4.6 & 52.4 ± 12.7 & 74.1 ± 12.2\\

 & \faCheck & \faCheck & \textbf{58.5} ± 0.9 & \textbf{73.0} ± 13.4 & \textbf{74.4} ± 1.0 & \textbf{75.8} ± 4.8 & \textbf{67.7} ± 14.7 & \textbf{95.7} ± 0.4 \\ \bottomrule
\end{tabular}%
}

\caption{Results comparing the individual contributions of the t-SNE~\cite{tsne} dimension reduction method and GCDformer for optimizing the GCD latent spaces. The use of each component is denoted as a \faCheck \space if it is used or a \faTimes \space if it is not used. We also report the standard deviation for these results. The usage of both t-SNE and GCDformer is the full OmniGCD method, as such these results are the same as Table 4 of the main paper. We \textbf{bold} the best results for each comparison.}
\label{tab:further_analysis}
\end{table}

\begin{table}[t]
\resizebox{\linewidth}{!}{%
\begin{tabular}{@{\extracolsep{4pt}}llccccccccccccccccccccc}
\toprule
 &  & \multicolumn{3}{c}{\faEye[regular] \space CIFAR-10} & \multicolumn{3}{c}{\faEye[regular] \space CIFAR-100} & \multicolumn{3}{c}{\faEye[regular] \space ImageNet-100} & \multicolumn{3}{c}{\faEye[regular] \space CUB-200} & \multicolumn{3}{c}{\faEye[regular] \space Stanford Car} & \multicolumn{3}{c}{\faEye[regular] \space Aircraft} & \multicolumn{3}{c}{\faEye[regular] \space Herbarium-19} \\ \cline{3-5}\cline{6-8}\cline{9-11}\cline{12-14}\cline{15-17}\cline{18-20}\cline{21-23}
\rule{0pt}{11pt} &  & All & Old & New & All & Old & New & All & Old & New & All & Old & New & All & Old & New & All & Old & New & All & Old & New \\ \midrule

\multirow{3}{*}{\rule{0pt}{2ex} \rotatebox{90}{\scriptsize DINOv1~\cite{dinov1}}} & PCA & 47.2 & 66.6 & 45.3 & 10.7 & 13.2 & 10.5 & 16.6 & 9.4 & 4.8 & 10.6 & 11.3 & 20.3 & 5.9 & 5.5 & 5.8 & 9.4 & 7.8 & 5.8 & 25.9 & 28.6 & 45.0 \\

 & UMAP & \textbf{91.4} & 96.4 & \textbf{94.0} & 57.6 & 82.6 & 74.3 & 77.2 & 89.0 & 77.0 & 41.3 & 60.0 & 25.1 & 10.4 & 5.5 & 11.9 & 16.2 & 11.9  & 11.2 & 27.7 & 31.4 & 45.0  \\
 & t-SNE & 90.7 & \textbf{97.0} & 91.6 & \textbf{60.0} & \textbf{89.9} & \textbf{74.6} & \textbf{81.1} & \textbf{91.0} & \textbf{74.8} & \textbf{44.5} & \textbf{66.0} & \textbf{33.9} & \textbf{12.6} & \textbf{19.6} & \textbf{16.9} & \textbf{18.9} & \textbf{14.3} & \textbf{13.9} & \textbf{30.8} & \textbf{48.6} & \textbf{50.0} \\
 \midrule
 
\multirow{3}{*}{\rule{0pt}{2ex} \rotatebox{90}{\scriptsize DINOv2~\cite{dinov2}}} & PCA & 70.3 & 64.5 & 84.0 & 12.7 & 9.0 & 2.1 & 20.6 & 7.8 & 6.8 & 16.9 & 27.7 & 12.1 & 13.1 & 21.2 & 16.7 & 16.0 & 22.7 & \textbf{12.2} & 26.1 & 13.9 & 9.7\\
& UMAP & \textbf{97.1} & \textbf{98.0} & 79.1 & 73.1 & \textbf{52.8} & 62.4 & 84.8 & \textbf{97.0} & \textbf{91.0} & 76.7 & \textbf{100.0} & \textbf{98.2} & 30.1 & \textbf{24.4} & 43.0 & 18.1 & 12.7 & 3.9 & 30.9 & 41.5 & 15.5\\
 & t-SNE & 96.9 & 96.9 & \textbf{95.6} & \textbf{78.1} & 47.9 & \textbf{56.8} & \textbf{88.7} & 94.0 & 90.0 & \textbf{79.8} & \textbf{100.0} & 96.4 & \textbf{33.4} & 24.2 & \textbf{43.1} & \textbf{21.2} & \textbf{17.0} & 11.6 & \textbf{34.8} & \textbf{60.0} & \textbf{25.8} \\

 \midrule
 &  & \multicolumn{3}{c}{\faFile*[regular] BANKING} & \multicolumn{3}{c}{\faFile*[regular] StackOverflow} & \multicolumn{3}{c}{\faFile*[regular] CLINIC} & \multicolumn{3}{c}{\faMicrophone \space VocalSet} & \multicolumn{3}{c}{\faMicrophone \space UrbanSound} & \multicolumn{3}{c}{\faSatellite \space EuroSAT} & \multicolumn{3}{c}{\faSatellite \space So2SAT~\cite{so2sat}} \\ \cline{3-5}\cline{6-8}\cline{9-11}\cline{12-14}\cline{15-17}\cline{18-20}\cline{21-23}
\rule{0pt}{11pt} & & All & Old & New & All & Old & New & All & Old & New & All & Old & New & All & Old & New & All & Old & New & All & Old & New \\ \midrule

 & PCA & 17.5 & 16.0 & 25.6 & 35.3 & 34.8 & 43.2 & 13.8 & 8.7 & 11.0 & 19.8 & 14.2 & 29.3 & 32.3 & 36.8 & 44.6 & 37.7 & 46.5 & 56.6 & 26.3 & 28.0 & 61.1    \\

 & UMAP & 65.4 & \textbf{86.0} & 22.8 & \textbf{87.6} & 75.0 & \textbf{74.2} & \textbf{82.9} & 45.2 & 68.1 & 21.1 & 14.4 & 29.4 & 41.2 & 21.8 & 52.6 & 55.1 & 68.7 & 46.7 & 30.8 & 31.5 & 37.9 \\
 & t-SNE & \textbf{66.4} & 77.1 & \textbf{40.8} & 86.8 & \textbf{78.0} & 73.8 & 82.1 & \textbf{96.0} & \textbf{90.0} & \textbf{25.6} & \textbf{17.5} & \textbf{49.2} & \textbf{45.5} & \textbf{56.7} & \textbf{63.9} & \textbf{68.3} & \textbf{75.5} & \textbf{69.5} & \textbf{31.7} & \textbf{38.4} & \textbf{58.4} \\
 
 \midrule
 &  & \multicolumn{3}{c}{\faSatellite \space RESIC45} & \multicolumn{3}{c}{\faSatellite \space UC Merced} \\ \cline{3-5}\cline{6-8}
\rule{0pt}{11pt}  & & All & Old & New & All & Old & New  \\ \cline{3-8}
\rule{0pt}{11pt} & PCA & 15.7 & 16.8 & 30.8 & 37.1 & 55.9 & 40.0\\
& UMAP & 56.1 & 67.8 & 69.3 & 73.0 & \textbf{76.5} & 74.5 \\
 & t-SNE & \textbf{58.5} & \textbf{73.0} & \textbf{74.4} & \textbf{75.8} & 67.7 & \textbf{95.7} \\

 \bottomrule
\end{tabular}%
}

\caption{Comparing OmniGCDs performance using PCA~\cite{pca}, UMAP~\cite{umap} and t-SNE~\cite{tsne} across all datasets and modalities. t-SNE achieves the best performance overall. We \textbf{bold} the best results for each comparison.}
\label{tab:additional_dim_reduction}
\end{table}
\end{landscape}

%% file: supp_sec/finetuning.tex
\section{Additional Results for Vision Encoder Fine-tuning}
\label{sec:vis_enc_ft}
Here, we provide additional results for Table~5 in the main paper, focusing on OmniGCD's performance with a fine-tuned DINOv1~\cite{dinov1} encoder. As noted in the main paper, this vision encoder was fine-tuned using GCD's~\cite{GCD2022} supervised and self-supervised contrastive training methods. The results in Table~\ref{tab:additional_FT} show that fine-tuning significantly improves performance on datasets where the base DINOv1 encoder performs poorly (CUB-200~\cite{cub200}, Stanford Cars~\cite{stanford_cars}, and Aircraft~\cite{aircraft}). However, these gains are not consistent across all datasets. While \textit{All} accuracy improves on CIFAR-10~\cite{cifar}, CIFAR-100~\cite{cifar}, and Herbarium-19~\cite{herb19}, fine-tuning yields lower \textit{All} accuracy on ImageNet-100. Although, for ImageNet-100, it does improve \textit{Old} and \textit{New} accuracy. Overall, these findings indicate that fine-tuning may be necessary when encoders fail to accurately encode the target data; future methods should explore how to best integrate the benefits of fine-tuning with modality-agnostic GCD.

%% file: supp_sec/dim_reduction_results.tex
\section{Dimension Reduction Options: Additional Results}
\label{sec:dim_reduction_method_abs}
In this section, we provide additional results for Table~8 in the main paper, comparing OmniGCD's performance across PCA~\cite{pca}, UMAP~\cite{umap}, and t-SNE~\cite{tsne} as dimensionality reduction methods. The results in Table~\ref{tab:additional_dim_reduction} show that t-SNE yields the best performance on 54 of the 69 metrics measured across all datasets and modalities. Notably, UMAP ranks second, with PCA performing significantly worse overall. These results further validate our choice of t-SNE for OmniGCD.